\RequirePackage{etoolbox}

\documentclass[ba,preprint]{imsart}
\pubyear{2019}

\usepackage{amsthm}
\usepackage{amsmath}
\usepackage{natbib}
\usepackage[colorlinks,citecolor=blue,urlcolor=blue,filecolor=blue,backref=page]{hyperref}
\usepackage{graphicx}

\RequirePackage[OT1]{fontenc}

\usepackage{amssymb, bm}
\usepackage{mathrsfs} 
\usepackage[capitalize,nameinlink]{cleveref}
\usepackage{mathtools}
\usepackage{thmtools}
\usepackage{enumitem}
\usepackage{subfigure}
\usepackage{color}

\def\[#1\]{\begin{align}\centering#1\end{align}}
\newcommand{\defas}{\vcentcolon=}  
\newcommand{\given}{\mid}

\newcommand{\event}[1]{(#1)}
\newcommand{\cevent}[1]{[#1]}

\newcommand{\Reals}{\mathbb{R}}

\newcommand{\dee}{\mathrm{d}}

\newcommand{\Nats}{\mathbb{N}}

\renewcommand{\Pr}{\mathbb{P}}

\def\EE{\mathbb{E}}

\newcommand{\Normal}{\mathcal{N}}
\newcommand{\dist}{\ \sim\ }

\newcommand{\bspace}{\Omega}
\newcommand{\bsa}{\mathcal A}
\newcommand{\borelspace}{(\bspace,\bsa)}

\newcommand{\PosReals}{\Reals_{>0}}

\newcommand{\betadist}{\mathrm{beta}}
\newcommand{\poisson}{\mathrm{Poisson}}
\newcommand{\gammadist}{\mathrm{gamma}}
\newcommand{\expdist}{\mathrm{exponential}}

\newcommand{\iid}{i.i.d.}

\newcommand{\nprocess}[3]{(#1_{#3})_{#3 \in #2}}
\newcommand{\process}[2]{\nprocess{#1}{#2}n}

\newcommand{\theset}[1]{\lbrace #1 \rbrace}

\newcommand{\BM}{B_0} 
\newcommand{\BMd}{b_0}

\newcommand{\bomega}{\omega}

\newcommand{\bb}{b}

\newcommand{\gprocess}[2]{(#1)_{#2}}

\newcommand{\card}[1]{\vert {#1} \vert}

\newcommand{\randfa}[1]{F_{#1}}

\newcommand{\totalmass}{\gamma}
\newcommand{\conc}{\theta}
\newcommand{\discount}{\alpha}

\newcommand{\randpartition}[1]{\Pi_{#1}}

\newcommand{\EPPF}{f_{\Pi}}

\newcommand{\GFC}{\mathscr C}
\newcommand{\tends}{\sim}
\newcommand{\levygbp}{\nu_{\Pi}}

\newcommand{\numblocks}{B}
\newcommand{\poststruct}{g}
\newcommand{\primitive}[3]{Q^{\, #1}_{\discount,\Theta}(#2,#3)}

\newcommand{\numfeat}{K}
\newcommand{\stable}{f_\discount}

\newcommand{\Varray}{\overline V}

\startlocaldefs
\numberwithin{equation}{section}
\theoremstyle{plain}

\theoremstyle{definition}

\endlocaldefs

\begin{document}

\begin{frontmatter}


\title{Gibbs-type Indian buffet processes}
\runtitle{Gibbs-type IBPs}

\begin{aug}
\author{\fnms{Creighton} \snm{Heaukulani}\thanksref{addr1,t1}\ead[label=e1]{c.k.heaukulani@gmail.com}}
\and
\author{\fnms{Daniel M.} \snm{Roy}\thanksref{addr2,t2}\ead[label=e2]{droy@utstat.toronto.edu}}

\runauthor{C. Heaukulani and D. M. Roy}

\address[addr1]{University of Cambridge, Cambridge, United Kingdom; 
	\printead{e1} 
}

\address[addr2]{University of Toronto, Toronto, Canada; 
	\printead{e2}
}

\end{aug}

\begin{abstract}
We investigate a class of feature allocation models that generalize the Indian buffet process and are parameterized by Gibbs-type random measures. 
Two existing classes are contained as special cases: the original two-parameter Indian buffet process, corresponding to the Dirichlet process, and the stable (or three-parameter) Indian buffet process, corresponding to the Pitman--Yor process.  
Asymptotic behavior of the Gibbs-type partitions, such as power laws holding for the number of latent clusters, translates into analogous characteristics for this class of Gibbs-type feature allocation models.
Despite containing several different distinct subclasses, the properties of Gibbs-type partitions allow us to develop a black-box procedure for posterior inference within any subclass of models.
Through numerical experiments, we compare and contrast a few of these subclasses and 
highlight the utility of varying power-law behaviors in the latent features.
\end{abstract}


\end{frontmatter}

\section{Introduction}

Feature allocation models \citep{GGS2007,broderick2013feature} assume that data are grouped into a collection of possibly overlapping subsets, called \emph{features}.
The best known example is the \emph{Indian buffet process} (IBP) \citep{GG06,GGS2007}, which has been successfully applied to a number of unsupervised learning problems in which the features represent unobserved/latent factors underlying the data.
While the IBP provides a nonparametric distribution suited to learning an appropriate number of features from the data, additional modeling flexibility---like heavy-tailed (i.e., power law) behavior in the number of latent features---is desirable in many applications.
Recent generalizations of the IBP addressing these needs parallel existing developments in the theory of random partitions.  
Indeed, random feature allocations may be viewed as a generalization of random partitions where the subsets of the partition are allowed to overlap.
In recent work, \citet{CUP} defines a broad class of random feature allocations called the \emph{generalized Indian buffet process}, each member of which corresponds to the law of an exchangeable partition.
In this article, we study the subclass corresponding to the random \emph{Gibbs-type partitions} \citep{gnedin2006gibbs}, which we call the \emph{Gibbs-type Indian buffet process} or simply \emph{Gibbs-type IBP}.
The Gibbs-type IBP inherits many useful properties from the Gibbs-type partitions (which include many of the partitioning models studied in the literature), and the special form of these models will allow us to develop practical black-box algorithms for simulation and posterior inference.

\subsection{Exchangeable feature allocations and the IBP}
\label{sec:ibp}

In the terminology introduced by \citet{broderick2013feature}, a \emph{feature allocation} of a set $A$ is a multiset of nonempty subsets of $A$, called \emph{features}, with the further restriction that no element of $A$ belongs to infinitely many features.
In statistical applications, we usually take $A$ to be the set $[n] \defas \{1, \dotsc, n\}$ for some $n \ge 1$, where $A$ then indexes a sequence of $n$ data points.
Intuitively, a random feature allocation of $[n]$ can be used to model $n$ data points in terms of latent features the data points share. 
Note that a data point may have multiple features, and so this notion generalizes clustering.

In many applications, we do not know the number of latent features necessary to adequately model a data set.
In such cases, one requires a nonparametric model in which the number of latent features is a random variable to be inferred from the data.
The canonical example of such a model is the Indian buffet process (IBP), introduced by \citet{GG06,GGS2007}.  An IBP is a random feature allocation with an a priori unbounded number of potential latent features
whose construction can be explained with the following culinary analogy:
Imagine a sequence of customers entering an Indian buffet restaurant. Each customer selects a finite number of dishes, chosen from a limitless supply of potential dishes to taste.
The first customer enters the buffet and takes $\poisson(\totalmass)$ dishes, where $\totalmass>0$ is called the \emph{mass parameter}.
For every $n\ge 1$, the $n+1$-st customer enters the buffet and decides to take each previously tasted dish $k$ with probability $n_k / (n + \conc)$, where $n_k$ is the number of previous customers that took dish $k$, and where $\conc>0$ is called the \emph{concentration parameter}.
The customer then takes $\poisson(\conc \totalmass/ (\conc+n))$ new (previously untasted) dishes.
For every $n\ge 1$, let $\numfeat_n$ denote the number of distinct dishes tasted among the first $n$ customers, and let $\randfa{n} \defas \{ \randfa{n,1}, \dotsc, \randfa{n,\numfeat_n} \}$, where $\randfa{n,1}, \randfa{n,2}, \dotsc$ are random subsets of $[n]$ such that $i \in \randfa{n,k}$ if and only if the $i$-th customer took the $k$-th dish, for every $i \le n$ and $k\le \numfeat_n$.
By construction, for every $n \ge 1$,
$\randfa{n}$ is a random feature allocation of $[n]$, 
and the sequence $\randfa{} \defas (\randfa{n})_{n\ge 1}$ defines a random feature allocation of $\Nats \defas \{1,2, \dotsc\}$.
We call $\randfa{}$ an \emph{Indian buffet process with mass parameter $\totalmass$ and concentration parameter $\conc$}.

\citet{GGS2007} show that,
for every $n\ge 1$, the distribution of $\randfa{n}$ is invariant to every permutation of $[n]$, i.e., the order of the customers does not influence the distribution of the resulting feature allocation.  A feature allocation with this property is called \emph{exchangeable} 
in analogy to exchangeable sequences, which satisfy a related family of distributional invariance properties.
Indeed, much of the recent work on exchangeable feature allocations has been inspired by analogous work in the theory of exchangeable partitions.
In statistical applications, random feature allocations have been applied to many of the same clustering problems as random partitions, where the extra flexibility of overlapping cluster assignments has often resulted in improved modeling power. 
%

\subsection{The Gibbs-type IBP}
\label{sec:introgibbsibp}

The Gibbs-type Indian buffet process, or Gibbs-type IBP, defines a class of exchangeable feature allocations that generalizes the IBP.
Let $\discount < 1$, which we will call the \emph{discount parameter}, and let ${\Varray \defas (V_{n,k} \colon n \ge k \ge 1)}$ be a triangular array of non-negative weights satisfying $V_{1,1} = 1$ and the recursive equations
\[
\label{eq:gibbsweights}
V_{n,k} = (n-\discount k) V_{n+1,k} + V_{n+1,k+1}
	,
	\qquad n \ge k \ge 1.
\]
(In the cases we will study, the weights $\Varray$ themselves are determined by a finite set of parameters, which we will denote by $\Theta$.)
Define the primitives
\[
\label{eq:primitives}
\primitive{n}{z_1}{z_2}
\defas
\sum_{k=1}^n
	\frac{ V_{n+z_1,k+z_2} }{ \discount^k } 
	\GFC(n,k;\discount)
	,
	\qquad n \ge z_1 \ge 1, \: z_2 \in \{0,1\}
	,
\]
where $\GFC(n,k; \discount)$ denotes the generalized factorial coefficient 
\[
\label{eq:gfc}
\GFC(n,k;\discount)
	\defas \frac 1 {k!} \sum_{i=0}^k (-1)^i \binom{k}{i} (-i\discount)_n
	,
	\qquad n \ge k \ge 1
	,
\]
and $(a)_n \defas \Gamma(a+n)/\Gamma(a)$. 
(See \citet{charalambides2005combinatorial} for a background on the generalized factorial coefficients.)
Then the Gibbs-type IBP may be described as follows:
Let $\totalmass > 0$, and imagine a sequence of customers entering an Indian buffet restaurant.
\begin{itemize}[noitemsep]
\item The first customer tries $\poisson(\totalmass)$ dishes from the buffet.

\item For every $n\ge 1$, the $n+1$-st customer

\begin{itemize}[noitemsep]
\item[--] tries each previously tasted dish $k$ independently with probability
$$
(S_{n,k}-\discount) \primitive{n}{1}{0},
$$
where $S_{n,k}$ is the number among the first $n$ customers that tried dish $k$;

\item[--] and tries
$ 
\poisson (
	\totalmass
	\primitive{n}{1}{1}
	 )
$
new dishes from the buffet.  

\end{itemize}
\end{itemize}
Construct a random feature allocation $\randfa{}$ of $\Nats$ from the actions of the customers, as described in \cref{sec:ibp}.
We call $\randfa{}$ a \emph{Gibbs-type Indian buffet process with parameters $(\totalmass, \discount, \Varray)$}. 
Like the original IBP, $\randfa{}$ is exchangeable, a property that will become clear in \cref{sec:gibbstypebp} when we provide an alternative construction via exchangeable sequences of random measures.

The reader familiar with the theory of Gibbs-type partitions (which we review in \cref{sec:partitions}) will recognize the recursive set of weights $\Varray$ appearing in \cref{eq:gibbsweights}, which along with the discount parameter $\discount$ determines the law of a \emph{Gibbs-type partition} \citep{gnedin2006gibbs}.
In what follows, we will see that every such choice $(\discount, \Varray)$ defining a subclass of the Gibbs-type partitions will determine a subclass of the Gibbs-type IBP.
In fact, some subclasses of Gibbs-type IBPs have already appeared in the literature, although they have not been presented from this perspective.
For example,
the \emph{stable} (or \emph{three-parameter}) IBP introduced by \citet{TG2009} and further studied by \citet{BJP2012}
is a Gibbs-type IBP
with the weights
 \[
\label{eq:PYweights}
V_{n,k} = \frac{ \prod_{\ell=1}^{k-1} ( \theta + \ell \discount ) }{ (\theta + 1)_{n-1} }
	,
	\qquad n \ge k \ge 1
	,
\]
for some parameter $\theta$ satisfying
\[
\label{eq:pyparams}
\begin{cases}
\theta > -\discount ,
	\quad &\text{when } \discount \in [0,1)
	,
	\\
\theta = m \vert \discount \vert \text{ for some } m \in \{1, 2, \dotsc\} ,
	\quad &\text{when } \discount<0 .
\end{cases}
\]
This setting of $(\discount, \Varray)$ corresponds to a subclass of the Gibbs-type partitions known as the \emph{two-parameter Chinese Restaurant processes}, i.e., the random partitions induced by the pattern of ties in exchangeable sequences sampled from a \emph{Pitman--Yor process} \citep{perman1992size,pitman1997two}.
(We will discuss the connection between exchangeable partitions and random probability measures in \cref{sec:partitions}.)
In this case, we have $\Theta = \{ \theta \}$ and the quantities $\primitive{n}{1}{0}$ and $\primitive{n}{1}{1}$ reduce to
\[
\primitive{n}{1}{0}
	= \frac{1}{\theta+n}
	\quad
	\text{and}
	\quad
\primitive{n}{1}{1}
	= \frac{\Gamma(\theta+1) \Gamma(\theta+\discount+n)}{\Gamma(\theta+n+1) \Gamma(\theta+\discount)}
	,
	\label{eq:PYIBPexplicit}
\]
respectively.
For $\alpha = 0$ and $\theta > 0$, we obtain the (two-parameter) IBP reviewed in \cref{sec:ibp}, and for $\alpha=0$ and $\theta=1$, the corresponding Gibbs-type IBP reduces to a more restrictive one-parameter variant of the IBP originally presented by \citet{GG06}.
In short, the stable IBP is the feature allocation analogue to the two-parameter Chinese Restaurant process, 
and the two-parameter IBP is the analogue to the one-parameter Chinese Restaurant process.

\subsection{Outline and summary of results}

In \cref{sec:partitions}, we review the theory of exchangeable Gibbs-type partitions, focusing on a few important subclasses.
In \cref{sec:gibbstypebp}, we derive the Gibbs-type IBP from a construction with completely random measures.
As an intermediate step, we define the \emph{Gibbs-type beta process}, a completely random measure that generalizes the \emph{beta process} introduced by \citet{Hjort1990}. 
We present stick-breaking constructions for the Gibbs-type beta process that generalize similar representations in the literature for the beta and stable beta processes \citep{TGG07,PZWGC2010,TG2009,BJP2012,carin2011variational}.
While these constructions are special cases of the \emph{generalized beta process} and corresponding \emph{generalized IBP} defined by \citet{CUP}, the special form of the Gibbs-type partitions will allow us to additionally derive practical algorithms for simulation and posterior inference with the Gibbs-type IBP.

Partitions with Gibbs-type structure exhibit many properties that are useful for applications. 
For example, when the discount parameter $\discount$ is in $(0,1)$, a Gibbs-type partition exhibits heavy-tailed (i.e., power law) behavior in the asymptotic distribution of the number of clusters induced by the partition. 
Latent features in the stable IBP were shown to exhibit analogous power-law behavior \citep{TG2009, BJP2012}, and in \cref{sec:powerlaw} we show that these characteristics are in a sense inherited from the two-parameter CRP or, equivalently, the Pitman--Yor process (with $\discount \in (0,1)$).
More generally, our results show that the Gibbs-type IBP inherits these power-law properties for any such class of partitioning models.
Similarly, when $\discount < 0$, the Gibbs-type partitions correspond to models with a random but finite number of clusters, and in \cref{sec:finitemodel} we show that the Gibbs-type IBP in this case corresponds to models with a random but finite number of features.

Many computations of interest with Gibbs-type partitions are expressed only through the parameters $(\discount, \Varray)$.
Likewise, the primitives $\primitive{n}{z_1}{z_2}$ in \cref{eq:primitives} only depend on these quantities.
Note that the description of the Gibbs-type IBP in the previous section only requires the arguments $\primitive{n}{1}{1}$ and $\primitive{n}{1}{0}$.
These quantities have probabilistic interpretations and are related to the well-studied probabilities of sampling a new and previous \emph{color} (or \emph{species}) under the law of a Gibbs-type partition.
(These concepts will be made clear in \cref{sec:partitions}.)
A likelihood function for the Gibbs-type IBP will be presented in \cref{sec:gibbstypebp}, which additionally requires the arguments $\primitive{n-s}{s}{1}$ for $s\le n$.
These terms also have probabilistic interpretations related to events in a Gibbs-type partition, which are all discussed in the supplementary material.
In \cref{sec:inference}, we derive a black-box posterior inference procedure that only requires these $n+1$ values of the primitives as input.
Finally, in \cref{sec:experiments} we demonstrate some of the practical differences between a few subclasses of the Gibbs-type IBP in a Bayesian nonparametric latent feature model applied to synthetic data and the classic MNIST digits dataset.

\section{Exchangeable Gibbs-type partitions}
\label{sec:partitions}

We briefly review the theory of Gibbs-type partitions; the reader should consult \citet{gnedin2006gibbs} for a more thorough treatment and \citet[Chs.~2~\&~3]{pitman2002combinatorial} for background on exchangeable partitions more generally.
Let $\randpartition{}$ be a random partition of $\Nats \defas \{1, 2, \dotsc\}$ into disjoint subsets, called \emph{blocks}.
We may write $\randpartition{} = \{A_1,A_2,\dotsc\}$, where $A_1$ is the block containing $1$ and $A_{k+1}$, for every $k \ge 1$, is the (possibly empty) block containing the least integer not in $A_1 \cup \dotsm \cup A_k$.
For every $n \ge 1$, let $\randpartition n$ be the restriction of $\randpartition{}$ to $[n] \defas \{1,\dotsc,n\}$.
For every $n \ge k \ge 1$, let $N_{n,k}$ be the number of elements in $A_{k} \cap [n]$, and let $\numblocks_n$ be the number of (nonempty) blocks in $\randpartition n$.
The partition $\randpartition{n}$ is said to be \emph{exchangeable} when its distribution is invariant under every permutation of the underlying set $[n]$ and $\randpartition{}$ is said to be exchangeable when every restriction $\randpartition{n}$, for $n \ge 1$, is exchangeable.%

The random partition $\randpartition{}$ is of \emph{Gibbs-type} when it is exchangeable and, 
for some $\discount < 1$ and $V_{n,k} \ge 0$, $n \ge k \ge 1$ satisfying \cref{eq:gibbsweights}, we have
\[
\EPPF ( n_1, \dotsc, n_k )
	&\defas
		\Pr \event{\numblocks_n = k, N_{n,1} = n_1, \dotsc, N_{n,k} = n_k } 
	\\
	&\: = V_{n,k} \prod_{\ell=1}^{k} (1-\discount)_{n_{\ell} - 1}
	,
	\label{eq:gibbseppf}
\]
for every $n \ge k \ge 1$ and $n_1, \dotsc, n_k \ge 1$ satisfying $\sum_j n_j = n$.
The function $\EPPF ( n_1,\dotsc,n_k )$, which is symmetric by exchangeability, is called the \emph{exchangeable partition probability function}, or EPPF.
The class of Gibbs-type partitions was introduced by \citet{gnedin2006gibbs} and has since been the subject of intense study due, in part, to the fact that the product form of the Gibbs-type EPPF in \cref{eq:gibbseppf} admits closed-form solutions for many quantities of interest. 

An exchangeable partition can be related to the pattern of colored balls drawn from an urn in a sequence of rounds as follows:
On each round, we may either (1) draw a ball from the urn at random, record the color, and place the ball back into the urn with another ball of the same color, or (2) we may place a ball of a new, previously unseen color into the urn.
The distinct colors of the balls correspond to the blocks in $\randpartition{}$, and the indices of the rounds during which a particular color was drawn indicates the members of the corresponding block.
In particular, on the first round the urn is empty and a ball of a new color is placed into the urn creating $\numblocks_1 = 1$ block.
We see from \cref{eq:gibbseppf} that during the $n+1$-st round, we draw a ball of the $k$'th previously seen color from the urn with probability
\[
\begin{split}
\Pr \cevent{ N_{n+1,k} > N_{n,k} | \numblocks_n, N_{n,1}, N_{n,2}, \dotsc } 
&= \frac{ \EPPF ( N_{n,1},\dotsc, N_{n,k} + 1, \dotsc ,N_{n,\numblocks_n} ) }{ \EPPF ( N_{n,1},\dotsc,N_{n,\numblocks_n}) }
	\\
&= \frac{ V_{n+1,\numblocks_n} }{ V_{n,\numblocks_n} }  (N_{n,k} - \discount)
		,
	\label{eq:probprevblock}
\end{split}
\]
for every $k \le \numblocks_n$, where $N_{n,k}$ denotes the size of the $k$-th block at the end of the $n$-th round. 
We draw a ball of a new color with probability
\[
\begin{split}
\Pr \cevent{ \numblocks_{n+1} > \numblocks_{n} | \numblocks_n, N_{n,1}, N_{n,2}, \dotsc } 
&= \frac{ \EPPF ( N_{n,1},\dotsc,N_{n,\numblocks_n}, 1 ) }{ \EPPF ( N_{n,1},\dotsc,N_{n,\numblocks_n}) } 
	\\
&= \frac{ V_{n+1,\numblocks_n+1} }{ V_{n,\numblocks_n} }
      .
      \label{eq:probnewblock}
\end{split}
\]
\citet[\textsection~2]{gnedin2006gibbs} show that the distribution of the number of blocks after the $n$'th round is given by
\[
\label{eq:blockdistr}
\Pr \event{ \numblocks_n = k }
	= \frac{V_{n,k}}{\discount^k} \GFC ( n, k; \discount )
	,
	\qquad k\le n
	,
\]
where $\GFC ( n, k; \discount )$ is the generalized factorial coefficient given in \cref{eq:gfc}.

The theory of exchangeable partitions is intimately connected to the theory of random probability measures.
In particular, by a representation theorem due to \citet{kingman1978representation}, every exchangeable partition may be obtained from the ties among an exchangeable sequence sampled from a random probability measure, and the laws of the partition and measure are one-to-one.
The measures inducing the Gibbs-type partitions are called \emph{Gibbs-type random measures}.
(The reader should consult \citet{kingman1975random} for background on random probability measures.)
We will focus on subclasses of the Gibbs-type partitions induced by several random probability measures that have been well-studied in the literature.
For example, the class of Gibbs-type random measures include the Dirichlet and Pitman--Yor processes already mentioned in the introduction.
Another subclass we will refer to frequently are those induced by the \emph{normalized generalized gamma processes} \citep{pitman2003poisson}, which have the weights
\[
V_{n,k} = \frac{ e^\beta \discount^{k-1} }{ \Gamma(n) }
	\sum_{i=0}^{n-1} \binom{n-1}{i} (-1)^i \beta^{i/\discount} 
		\Gamma ( k - i/\discount ; \beta )
	,
	\label{eq:NGGweights}
\]
where $\discount \in (0,1)$, $\beta>0$, and $\Gamma(x;a)\defas \int_x^\infty s^{a-1} e^{-s} \dee s$ is the incomplete gamma function. 
Special cases include the partitions induced by the \emph{normalized inverse Gaussian processes} \citep{lijoi2005hierarchical} when $\discount = 1/2$; the \emph{normalized $\discount$-stable processes} \citep{kingman1975random} in the limit $\beta\to 0$; and the Dirichlet processes, again, in the limit $\discount \to 0$.

More generally, \citet[Thm.~12]{gnedin2006gibbs} showed that the law of every Gibbs-type partition with fixed discount parameter $\discount<1$ is a unique probability mixture over one of three classes of extreme partitions, depending on the value of $\discount$.
When $\discount\in(0,1)$, the extreme partitions are induced by the \emph{Poisson--Kingman random measures} \citep{pitman2003poisson}; in this case, it follows from \citet[Prop.~9]{pitman2003poisson}, that
\[
\label{eq:PKweights}
V_{n,k} = \frac{\discount^k}{\Gamma(n-k\discount) \tau^{k\discount} \stable(\tau)}
		\int_0^1 p^{n-k\discount-1} \stable(\tau (1-p))  \dee p
	,
\]
for a parameter $\tau > 0$, where $\stable$ is the density of a positive $\discount$-stable random variable. 
Members of this subclass are obtained by mixing over $\tau$ with respect to a probability distribution on the positive real numbers.
Particular attention has been paid to the cases when $\tau$ has density function $h(t) \stable(t)$, for some measureable function $h \colon \PosReals \to \PosReals$.
For example, when $h(t) = \frac{\Gamma(\theta+1)}{\Gamma(\theta/\discount+1)} t^{-\theta}$ for some $\theta>-\discount$, then \cref{eq:PKweights} reduces to \cref{eq:PYweights} and we obtain the partitions induced by the Pitman--Yor processes (with $\discount \in (0,1)$).
When $h(t) = e^{\beta^\discount - \beta t}$ for some $\beta>0$, then \cref{eq:PKweights} reduces to \cref{eq:NGGweights} and we obtain the partitions induced by the normalized generalized gamma processes.
See \citet[\textsection5]{pitman2003poisson} for a further treatment.
When $\discount=0$, the extreme partitions are induced by the Dirichlet processes with concentration parameter $\theta>0$.
Members of the subclass are obtained by mixing over a random $\theta$ with respect to a probability distribution on $\PosReals$.
Finally, when $\discount<0$, the extreme partitions are induced by the Pitman--Yor processes with concentration parameter $\theta = m \card{\discount}$, for some positive integer $m$.
In this case, the weights in \cref{eq:PYweights} may be rewritten as
\[
V_{n,k}
	= \frac{ \card{\discount}^{k-1} \prod_{\ell=1}^{k-1} (m-\ell) }
	{ (m \card{\discount} + 1)_{n-1} }
	1_{\{1, \dotsc, \min(n,m)\}} (k)
	,
	\qquad
	n \ge k \ge 1 ,
\]
highlighting the restriction on the weights $\Varray$ to be non-negative.
This is equivalent to an urn scheme with a finite number $m$ of different colors \citep[Chp.~3, Sec.~2]{pitman2002combinatorial}.
Members of the subclass are obtained by mixing over a random $m$ with respect to a probability distribution on the positive integers.

In summary, each Gibbs-type partition with fixed $\discount < 1$ is a unique probability mixture over the extreme partitions induced by either:
\begin{enumerate}[noitemsep]
\item The Pitman--Yor processes with discount parameter $\discount$ and concentration parameter $\theta = m \card{\discount}$ for $m$ in $\Nats$, when $\discount < 0$;

\item The Dirichlet processes with concentration parameter $\theta>0$, when $\discount=0$; or

\item The Poisson--Kingman processes with parameter $\tau > 0$, when $\discount\in(0,1)$.
\end{enumerate}
It should be clear that each Gibbs-type partition defines a Gibbs-type IBP via the construction in \cref{sec:introgibbsibp}, and so it will suffice to characterize the Gibbs-type IBP in each of these regimes.
In order to perform posterior inference within each subclass of the Gibbs-type IBP, we will place prior distributions on any parameters defining the weights $\Varray$ and infer their values from data (see \cref{sec:inference}).

\section{Constructions from random measures}
\label{sec:gibbstypebp}

\citet{TJ2007} connected exchangeable feature allocations with the theory of completely random measures by showing that the IBP captures the combinatorial structure of an exchangeable sequence of Bernoulli processes directed by a beta process \citep{Hjort1990}.  
Generalizations of this approach have appeared in the literature, which, from our perspective, include generalizations of the IBP that are parameterized by the law of the Pitman--Yor processes \citep{TG2009,BJP2012} and, more generally, by the law of any exchangeable partition \citep{CUP}.
Here we describe the case corresponding to the Gibbs-type partitions.

\subsection{Gibbs-type beta processes}
\label{sec:bepbp}

Let $\randpartition{}$ be the exchangeable Gibbs-type partition defined by \cref{eq:gibbseppf}, whose restriction $\randpartition{n}$ to $[n]$ has block sizes (in order of appearance; see \cref{sec:partitions}) denoted by $N_{n,1}, N_{n,2}, \dotsc$, for every $n\ge 1$.
By Kingman's paint-box construction \citep{kingman1978representation},
the limiting relative frequencies of the blocks
\[
P_k \defas
	\lim_{n\to \infty} \frac{N_{n,k}}{n}
	\label{eq:limitfreq}
\]
exist almost surely for every $k \in \Nats$.  
Let $\mu_1$ be the distribution of $P_1$, which is called the \emph{structural distribution}.
The structural distribution reveals quite a bit about the exchangeable partition, but does not necessarily characterize it \citep[Chp.~2.3]{pitman2002combinatorial}; \citep{pitman1995exchangeable}.  The structural distribution will entirely determine the law of the corresponding Gibbs-type IBP.
Let $\bspace$ be a complete, separable metric space and let $\bsa$ be its Borel $\sigma$-algebra.
Following \citet[Thm.~1.2]{CUP}, define a purely atomic random measure $B$ on $\borelspace$ by
\[
\label{eq:gibbsbetaprocess}
B \defas \sum_{k \ge 1} \tilde \bb_k \delta_{\tilde \bomega_k}
	,
\]
where $(\tilde \bomega_1, \tilde\bb_1), (\tilde \bomega_2, \tilde\bb_2), \dotsc$ are the points of a Poisson process on $\bspace \times (0,1]$ with ($\sigma$-finite) intensity measure
\[
\label{eq:levygbp}
\levygbp (\dee \omega \times \dee p) \defas
	\BM (\dee \omega)\,
	 p^{-1} \mu_1 (\dee p)
		,
\]
for some non-atomic $\sigma$-finite measure $\BM$ on $\borelspace$.
Note that, because $\levygbp$ is not a finite measure, $B$ will have a countably infinite number of atoms, almost surely.
We call $B$ a \emph{Gibbs-type beta process with EPPF $\EPPF$ and base measure $\BM$}. 
Also note that the construction of $B$ ensures that the random variables $B(A_1),\dotsc,B(A_k)$ are independent for every finite, disjoint collection $A_1, \dotsc, A_k \in \bsa$, and $B$ is therefore said to be \emph{completely random} or have \emph{independent increments}.  (See \citet[Chp.~12]{Kallenberg2002} for a background on completely random measures.) 
Following \citet{TJ2007}, define a sequence $\process Z \Nats \defas (Z_1, Z_2, \dotsc)$ of random measures on $\borelspace$ that are conditionally \iid, given $B$, with
\[
\label{eq:gibbsbernoulliprocess}
Z_n 
	= \sum_{k\ge 1} 
		1_{\{ U_{n,k} < \tilde\bb_k \}}
		\delta_{\tilde \bomega_k}
	,
	\qquad 
	n \in \Nats
	,
\]
where $\gprocess{U_{n,k}}{n,k\in \Nats}$ is an independent collection of \iid\ $\text{Uniform}(0,1)$ random variables.
Then $\process Z \Nats$ is an exchangeable sequence of Bernoulli processes. 
By construction,
because $B$ is completely random, the elements of $\process Z \Nats$ are completely random, both conditionally on $B$, and unconditionally.

\newcommand{\ii}{j}
Fix $n\ge 1$.
We now describe the conditional distribution of $Z_{n+1}$ given $Z_{[n]} \defas (Z_1,\dotsc, Z_n)$.
A rigorous derivation can be found in \citet{CUP} and \citet[][Prop.~3.1]{james2017poisson}. The following exposition emphasizes intuition, and follows the approach of \citet{TG2009} and \citet{TJ2007}, who built on the work of \citet[][Thm.~3.3]{Kim1999}.
By the complete randomness of $Z_{n+1}$, we may first analyze the conditional distribution of its \emph{fixed atoms} (that is, any atoms that have also appeared among $Z_{[n]}$), followed by the conditional distribution of its \emph{ordinary component} (which consists of atoms that have not appeared among $Z_{[n]}$).
Consider the fixed atoms:
That is, let $(\bomega_1, \dotsc, \bomega_{\numfeat_n})$ be the $\numfeat_n$ distinct atoms among $Z_{[n]}$, 
listed in order of appearance (i.e., the order in which they first appear in the sequence, with ties broken uniformly at random and independently).  We can relate these distinct atoms to the atoms in $B$: 
we have $(\bomega_1,\dotsc,\bomega_{\numfeat_n}) = (\tilde \bomega_{\ii_1},\dotsc,\tilde \bomega_{\ii_{\numfeat_n}})$ for some random integers $(\ii_1,\dotsc,\ii_{K_n})$.
For $k\le \numfeat_n$, the measure $Z_{n+1}$ takes atom $\bomega_k$ with some probability $\bb_k = \tilde \bb_{\ii_k}$,
and the conditional distribution of $\bb_k$, given $Z_{[n]}$, is
\[
\Pr \cevent{ \bb_k \in \dee p \given Z_1, \dotsc, Z_n }
	&=
	\frac{ {p}^{S_{n,k}-1} (1-p)^{n-S_{n,k}} \mu_1(\dee p) }
		{ \poststruct(n,S_{n,k}) }
		,
\]
where $S_{n,k} \defas \sum_{j=1}^n Z_j (\theset{\bomega_k})$, for $k\le \numfeat_n$, and 
\[
\poststruct(n,s) \defas \int_{(0,1]} p^{s-1} (1-p)^{n-s} \mu_1(\dee p)
	,
	\qquad
	n \ge s \ge 1
	.
	\label{eq:integraldef}
\]
Therefore, for every $k \le \numfeat_n$, we have
\[ 
\Pr \cevent{Z_{n+1} (\theset{ \bomega_k }) =1 \given Z_1,\dotsc,Z_n }
	&= \EE [ \bb_k \given Z_1, \dotsc, Z_n ]
	\\
	&=
	\frac{ \poststruct(n+1,S_{n,k}+1) }{ \poststruct(n,S_{n,k}) }
	.
	\label{eq:fixedZ}
\]
Now consider the ordinary component:
Informally speaking, the distribution of the atoms of $Z_{n+1}$ that have not appeared among $Z_{[n]}$ may be described as follows: 
for some infinitesimal set ${\dee \omega \subseteq \bspace \setminus \{\bomega_1,\dotsc,\bomega_{K_n}\}}$,
\[
\label{eq:ordinaryZ}
\Pr \cevent{ Z_{n+1} (\dee \omega) = 1 \given Z_1,\dotsc,Z_n }
	&= \int_{(0,1]} p (1-p)^n \levygbp(\dee \omega \times \dee p)
	\\*
	&= \BM (\dee \omega) \poststruct(n+1,1) 
	.
\]
More precisely, on $\bspace \setminus \{\bomega_1,\dotsc,\bomega_{\numfeat_n}\}$, the measure $Z_{n+1}$ is a Poisson process with intensity measure ${\poststruct(n+1,1) \BM}$, and the number of new atoms in $Z_{n+1}$ is therefore Poisson distributed with rate $\totalmass \poststruct(n+1,1)$, where $\totalmass \defas \BM(\bspace)<\infty$.

\subsection{Exchangeable feature allocations of Gibbs-type}
\label{sec:gibbstypeibp}

We now construct an exchangeable feature allocation from the exchangeable sequence $\process Z \Nats$.
Recall the buffet process analogy introduced in \cref{sec:ibp,sec:introgibbsibp}.
Let $n\ge 1$.
For every $i\le n$, associate the Bernoulli process $Z_i$ with the $i$-th customer entering the Indian buffet restaurant, and associate the $\numfeat_n$ distinct atoms $(\bomega_1, \dotsc, \bomega_{\numfeat_n} )$ among $Z_{[n]}$ with the distinct dishes sampled among the first $n$ customers, where the dishes are listed in order of appearance, as described earlier.
Then $\numfeat_n$ represents the total number of dishes taken by the first $n$ customers, and $S_{n,k}$ is the number of customers, among the first $n$ customers, that sampled dish $k$. 
Let $\randfa{n, 1}, \randfa{n, 2},\dotsc$ be random subsets of $[n]$ such that, for every $i\le n$ and $k \le \numfeat_n$, we have $i \in \randfa{n,k}$ if and only if $Z_i (\theset{\bomega_k}) = 1$.
It is easy to verify that $\randfa{n} \defas \{ \randfa{n,1}, \dotsc, \randfa{n,\numfeat_n} \}$ is
a random feature allocation of $[n]$,
and $\randfa{} \defas (\randfa{n})_{n\in\Nats}$ is a random feature allocation of $\Nats$.
Because the sequence $\process Z \Nats$ is exchangeable, it follows that $\randfa{}$ is an exchangeable feature allocation of $\Nats$.
Note that $\randfa{}$ captures only the \emph{combinatorial structure} of the sequence $\process Z \Nats$---that is, the pattern of shared atoms among the elements of the sequence $\process Z \Nats$, ignoring the locations of the atoms themselves---analogously to the way exchangeable partitions only capture the combinatorial structure of exchangeable sequences of random variables directed by a random probability measure.

We will now show that the law of $\process Z \Nats$, and therefore the induced feature allocation $\randfa{}$, is characterized by the Gibbs-type IBP presented in \cref{sec:introgibbsibp}.
In particular, we will show the probability the $n+1$-st customer takes a previously sampled dish 
agrees with the probability that each atom in $Z_{[n]}$ appears in $Z_{n+1}$ (in \cref{eq:fixedZ}),
and the mean of the Poisson distributed number of new dishes taken by the $n+1$-st customer agrees with the mean of the Poisson distributed number of new atoms in $Z_{n+1}$.
To that end, it suffices to study the triangular array of integrals $g(n,s)$, for $n\ge s \ge 1$. 
The structural distribution $\mu_1$ relates the Gibbs-type beta process $B$ to the probabilities of combinatorial events in the exchangeable partition $\randpartition{n}$.
In particular, we have
\[
\poststruct(n,s)
	= \Pr \event{ N_{s,1} = s \, \cap \, N_{n,1} = s }
	= \Pr \event{ N_{n, \numblocks_{n-s+1}} = s }
	.
	\label{eq:structspecies}
\]
To understand these identities, we return to the urn scheme interpretation, described in \cref{sec:partitions}, which we recall is initialized by placing a colored ball into the urn. 
From \cref{eq:limitfreq}, we may informally interpret the structural distribution $\mu_1$ as the (asymptotic) probability of drawing this color from the urn in subsequent rounds of the scheme.
We may therefore interpret the definition of $g(n,s)$ in \cref{eq:integraldef} as the probability of drawing this color in the first $s$ rounds of the urn scheme, followed by not drawing it again in the following $n-s$ rounds, resulting in the first equality in \cref{eq:structspecies}.
The second equality follows by exchangeability, i.e., we may reorder the first $s$ draws from the urn scheme to instead be the last $s$ draws without affecting this probability.
Formal derivations of such formulae can be obtained with properties of the structural distribution, as discussed by \citet{pitman1995exchangeable}; \citet[\S~2.3]{pitman2002combinatorial}.

Clearly ${\poststruct(1,1)=1}$.
Consider $\poststruct(n+1,1) = \Pr \event{N_{n+1, \numblocks_{n+1}} = 1 } = \Pr \event{ \numblocks_{n+1} > \numblocks_{n} }$.
This is the probability that a new color is drawn on the $n+1$-st round, which conditioned on $\numblocks_n$ is given by $V_{n+1,\numblocks_n+1} / V_{n,\numblocks_n}$ (see \cref{eq:probnewblock}).
Then by taking an expectation over $\numblocks_n$ (with respect to \cref{eq:blockdistr}), we have for every $n \ge 1$,
\[
\Pr \event{ \numblocks_{n+1} > \numblocks_n }
	= \EE \biggl [
		\frac{ V_{n+1,\numblocks_n+1} }{ V_{n,\numblocks_n} }
		\biggr ]
	= \sum_{k=1}^n
			\Bigl(
			\frac{ V_{n+1,k+1} }{ \discount^k }
			\GFC(n,k;\discount)
			\Bigr )
	= 
	\primitive{n}{1}{1}
	, \nonumber
\]
where we recall that $\primitive{n}{\cdot}{\cdot}$ was given by \cref{eq:primitives}.
This is the mean number of new dishes tasted by the $n+1$-st customer in the Gibbs-type IBP, as desired.
In general,
$
\poststruct(n,s) 
	= \Pr \{ N_{n, \numblocks_{n-s+1}} = s \}
$
is the probability that a new color is drawn on the $(n-s+1)$-st iteration and then drawn again $s-1$ times in a row.
Conditioned on $\numblocks_{n-s}$, sampling a new color occurs with probability $V_{n-s+1,\numblocks_{n-s}+1} / V_{n-s,\numblocks_{n-s}}$, and drawing this color $s-1$ additional times occurs with probability
\[
\label{eq:generalg}
\begin{split}
&\frac{ V_{n-s+2,\numblocks_{n-s}+1} }{ V_{n-s+1,\numblocks_{n-s}+1} } (1-\discount)
	\frac{ V_{n-s+3,\numblocks_{n-s}+1} }{ V_{n-s+2,\numblocks_{n-s}+1} } (2-\discount)
	\cdots \frac{ V_{n,\numblocks_{n-s}+1} }{ V_{n-1,\numblocks_{n-s}+1} } (s-1-\discount)
	\\
	&\qquad
	= \frac{ V_{n,\numblocks_{n-s}+1} }{ V_{n-s+1,\numblocks_{n-s}+1} }
		(1-\discount)_{s-1}
	.
\end{split}
\]
Multiplying, we have
\[
\Pr \cevent{ N_{n, \numblocks_{n-s+1}} = s \given \numblocks_{n-s} }
	= \frac{ V_{n,B_{n-s}+1} }{ V_{n-s,B_{n-s}} } (1-\discount)_{s-1}
	.
	\label{eq:condint}
\]
With an iterated expectation and the form of \cref{eq:condint}, we may write
\[
\frac{\poststruct(n+1,s+1)}{\poststruct(n,s)}
	&= \EE \biggl [
		\frac{ \Pr \cevent{ N_{n+1, \numblocks_{n-s+1}} = s+1 \given \numblocks_{n-s} } }
			{ \Pr \cevent{ N_{n, \numblocks_{n-s+1}} = s \given \numblocks_{n-s} } }
		\biggr ]
		\\
	&
	= (s-\discount)
		\EE \biggl [
		\frac{ V_{n+1,\numblocks_{n-s} + 1} }{ V_{n,\numblocks_{n-s}+1} }
		\biggr ]
	.
	\label{eq:gratio}
\]
Recall that, on the event $\{ N_{n,\numblocks_{n-s+1}} = s \}$, we have $\numblocks_{n-s}+1 = \numblocks_{n-s+1} = \numblocks_n$. Therefore,
\[
(s-\discount)
		\EE \biggl [
		\frac{ V_{n+1,\numblocks_n} }{ V_{n,\numblocks_n} }
		\biggr ]
	= (s-\discount) \sum_{k=1}^n 
			\frac{ V_{n+1,k} }{ \discount^k } \GFC(n,k;\discount) 
		= (s-\discount) \primitive{n}{1}{0}
			,
		\label{eq:prevdishderiv}
\]
which shows that \cref{eq:fixedZ} is the probability the $n+1$-st customer in the Gibbs-type IBP tastes a dish that has been tasted $s$ times previously.
Finally, let $\BMd$ denote a probability density function for the normalized base measure $\gamma^{-1} \BM$.
In the supplementary material, we show that $(Z_1, \dotsc, Z_n)$ has a probability density function $p_n$
given by
\[
\label{eq:pmf}
\begin{split}
p_n( Z_1, \dotsc, Z_n )
	&=
		\totalmass^{\numfeat_n}
			 \exp \Bigl (
			 	 -\totalmass 
				 \sum_{j=1}^n \primitive{j-1}{1}{1} 
				\Bigr )
				\\
		&\qquad \times		
		\prod_{k=1}^{\numfeat_n}
			\bigl [
			 (1-\discount)_{S_{n,k}-1} \primitive{n-S_{n,k}}{S_{n,k}}{1}
		\BMd ( \bomega_k )
		\bigr ]
	,
	\end{split}
\]
where $\primitive{0}{n}{1} \defas (1-\discount)_{n-1} V_{n,1}$ for every $n\ge 1$.

\subsection{Special cases}
\label{sec:specialcases}

Clearly, any EPPF of the Gibbs-type form in \cref{eq:gibbseppf} will induce a Gibbs-type IBP.
Some special cases of these constructions are already known in the literature.
We have already discussed the Gibbs-type IBPs corresponding to partitions induced by the Pitman--Yor (and, thus, Dirichlet) processes.
Indeed, in the Pitman--Yor process case the structural distribution is $\mu_1 = \betadist(1-\discount, \theta+\discount)$ for $\discount \in [0,1)$ and $\theta>-\discount$.
In this case, the Gibbs-type beta process specializes to the \emph{stable} (or \emph{three-parameter}) \emph{beta process} \citep{TG2009}, which contains the original beta process when $\discount=0$.
Despite those authors not studying the case when $\discount < 0$ and $\theta= m\card{\discount}$, for some $m$ in $\Nats$, we may just as well define this extension of the stable beta process and stable IBP, and, indeed, the structural distribution (and so the construction of $B$ and $\process Z \Nats$) are of the same form.
See \citep[Prop.~9 and the text following]{pitman1995exchangeable} for references on the structural distributions in all of these cases.

As described at the end of \cref{sec:partitions}, the only remaining case of the Gibbs-type IBPs to consider are those corresponding to the Gibbs-type partitions with $\discount \in (0,1)$, which are the partitions induced by the Poisson--Kingman processes with parameter $\tau>0$.
In this case, \citet[Sec.~5.4]{pitman2003poisson} shows that the structural distribution $\mu_1$ admits a probability density function on $(0,1)$ given by
\[
\label{eq:PKstruct}
p(v)
	= \frac{ \discount }{ \Gamma(1-\discount) }
		v^{-\discount}
		\tau^{-\discount}
		\frac{ \stable(\tau (1-v)) }{ \stable(\tau) }
	,
\]
which was also derived by \citet{favaro2013slice} with an application of \citet[Thm.~2.1]{perman1992size}.
For the remainder, we will refer to any subclass of the Gibbs-type beta process or IBP by the name of the random measures inducing the corresponding Gibbs-type partition.
For example, we will say \emph{Pitman--Yor-type beta process} and \emph{Pitman--Yor-type IBP} instead of stable beta process and stable IBP, etc.

\section{Stick-breaking representations}
\label{sec:stickbreaking}

So-called \emph{stick-breaking} representations for the beta process are analogous to the stick-breaking constructions for random probability measures such as the Dirichlet and Pitman--Yor processes.  
(See \cite{sethuraman1994stick,ishwaran2001gibbs} for background on stick-breaking representations for random probability measures.)
These representations are useful for applications because they lead to practical inference procedures.
With an application of \citet[Thm.~1.3]{CUP}, we may obtain an analogous stick-breaking representation of the Gibbs-type beta process as follows: 
Recall from \cref{eq:limitfreq} that $P_i$ is the limiting frequency of the $i$-th block in a Gibbs-type partition, whose distribution is denoted by $\mu_i$.  Then $(P_i)_{i\in \Nats}$ are the size-biased frequencies.
Let $\BM$ be a non-atomic measure on $\borelspace$, and define 
\[
B \defas \sum_{i=1}^\infty
	\sum_{j=1}^{C_i}
		P_{i,j}
		\delta_{\omega_{i,j}}
	,
	\label{eq:stickbreakingrep}
\]
where $\gprocess{C_i}{i\in\Nats}$, $\gprocess{\omega_{i,j}}{i,j\in\Nats}$, and $\gprocess{P_{i,j}}{j\in\Nats}$ are independent processes and
\begin{enumerate}[noitemsep]
\item $\gprocess{C_i}{i\in\Nats}$ are \iid\ $\poisson(\totalmass)$ random variables with $\totalmass \defas \BM(\bspace)$;

\item $\gprocess{\omega_{i,j}}{i,j\in\Nats}$ are \iid\ random elements in $\bspace$ with distribution $\totalmass^{-1} \BM$; and

\item For every $i\in\Nats$, the random variables in the collection $\gprocess{P_{i,j}}{j\in\Nats}$ are \iid\ copies of $P_i$.

\end{enumerate}

The problem of constructing $B$ then amounts to that of constructing the size-biased frequencies $\gprocess{P_i}{i\in\Nats}$ specific to the underlying Gibbs-type partition.
Efficient constructions for these size-biased frequencies are available for many subclasses of the Gibbs-type partitions; in these cases, we obtain efficient stick-breaking constructions for the corresponding subclasses of the Gibbs-type beta process.
Here we summarize these results.

For every $i\in\Nats$, let
\[
\label{eq:stickdistr}
P_i = 
W_i \prod_{j=1}^{i-1} (1-W_j )
	,
\]
with $P_1 = W_1$, for some random elements $W \defas \gprocess{W_j}{j\in\Nats}$ in $(0,1]$.
If $W_j \dist \betadist(1,\theta)$, \iid\ for every $j\in\Nats$ and $\theta>0$, then \cref{eq:stickdistr} is the $i$-th stick of a Dirichlet process \citep{sethuraman1994stick}.
In our terminology, \citet{PZWGC2010} showed that $B$ is then a Dirichlet-type beta process (with concentration parameter $\theta$ and base measure $\BM$).
If the random variables $W$ are merely independent with $W_j \dist \betadist(1-\discount, \theta + j \discount)$, for every $j \in \Nats$ and some $\discount \in (0,1)$ and $\theta > -\discount$, then \cref{eq:stickdistr} is the $i$-th stick of a Pitman--Yor process \citep{perman1992size}.
In our terminology, \citet{BJP2012} showed that $B$ is a Pitman--Yor-type beta process (with discount parameter $\discount$, concentration parameter $\theta$, and base measure $\BM$).
As with the Pitman--Yor-type IBP, these authors did not consider a stick-breaking construction for the Pitman--Yor-type beta process with $\discount<0$ and $\theta=m\card{\discount}$ for some $m$ in $\Nats$.
However, the sticks of the Pitman--Yor processes in this case are still independent and distributed as $W_j \dist \betadist(1-\discount, m\vert\discount\vert + j \discount)$, for every $j \in \Nats$ \citep[Prop.~9]{pitman1995exchangeable}, and so this extension does indeed arise from the construction in \cref{eq:stickbreakingrep}.

In order to complete the stick-breaking representations for the Gibbs-type beta processes, all that remains is to describe the distribution of $W$ in the case when $\discount\in(0,1)$.
\citet{favaro2013slice} applied \citep[Thm.~2.1]{perman1992size} to show that the sequence $\gprocess{W_j}{j\in\Nats}$ is composed of dependent random variables that may be characterized sequentially as follows:
The first stick $P_1 = W_1$ has distribution $\mu_1$ given by \cref{eq:PKstruct}.
For every $j\ge 2$, conditioned on $W_1, \dotsc, W_{j-1}$, the random variable $W_j$ admits a conditional density on $(0,1]$ with density function
\[
\begin{split}
p( w_j \given w_1, \dotsc, w_{j-1} )
	&=
	\frac{\discount}{\Gamma(1-\discount)}
		\Bigl [
			\tau w_j \prod_{k=1}^{j-1} (1-w_k) 
		\Bigr ]^{-\discount}
		\frac{ f_\discount(\tau \prod_{k=1}^j (1-w_k))}{ f_\discount(\tau \prod_{k=1}^{j-1} (1-w_k)) }
	,
\end{split}
\]
where $\tau>0$ is the parameter of the Poisson--Kingman model (see \cref{eq:PKweights}).
An algorithm for slice sampling the sequence $W$ was provided therein, and \citet{favaro2014stickbreaking} showed that, under certain assumptions on the parameter $\discount$, these sticks can be directly constructed with beta and gamma random variables.

We present an alternative stick-breaking representation for the Gibbs-type beta process that represents the measures $\sum_{j=1}^{C_i} P_{i,j} \delta_{\omega_{i,j}}$, for every $i\ge 1$, in \cref{eq:stickbreakingrep} with independent Poisson processes.
This representation results from an application of \citet[Thm.~1.4]{CUP}. 
Let
\[
B = \sum_{n=0}^\infty
	\sum_{(\omega,p) \in \eta_n}
		p\, \delta_\omega
	,
	\label{eq:stickbreakingrep2}
\]
where $\eta_0, \eta_1, \eta_2, \dotsc$ are independent Poisson processes on $\bspace \times (0,1]$ with finite intensity measures
\[
\label{eq:stickbreakingrep2pp}
(\EE \eta_n) (\dee \omega \times \dee p) = \BM( \dee \omega ) (1-p)^{n} \mu_1(\dee p)
	,
	\qquad
	n\in\{0, 1, 2, \dotsc \}
	.
\]
One may verify that $B$ in \cref{eq:stickbreakingrep2} is indeed the Gibbs-type beta process given by \cref{eq:levygbp}
using a Poisson process superposition argument and the identity $p^{-1} = \sum_{n=0}^\infty (1-p)^{n}$.
For $\discount<0$, we have
\[
\label{eq:pybpstickpp}
\begin{split}
(\EE \eta_n) ( \dee \omega \times \dee p )
	&= \BM ( \dee \omega )
		\frac{\Gamma(1+\theta)}{\Gamma(1-\discount) \Gamma(\theta+\discount)}
			p^{-\discount} (1-p)^{\theta+\discount+n-1}			
			\dee p
		,
\end{split}
\]
where $\theta = m \card{\discount}$ for some $m$ in $\Nats$.
This same form characterizes the case $\discount=0$ by setting $\theta>0$.
When $\discount \in (0,1)$ and $\theta>-\discount$, \cref{eq:pybpstickpp} characterizes the rest of the Pitman--Yor-type beta processes. 
More generally, when $\discount \in (0,1)$ we have
\[
\begin{split}
(\EE \eta_n) ( \dee \omega \times \dee p )
	&= \BM ( \dee \omega )
		\frac{\discount}{\Gamma(1-\discount)}
			(1-p)^{n}
			p^{-\discount}
			\tau^{-\discount} \frac{ \stable(\tau (1-p)) }{ \stable(\tau) }
			\dee p
		,
\end{split}
\]
where $\tau>0$.

These stick-breaking representations are useful for applications because inference procedures may be obtained in which the sticks are auxiliary variables.  
Though only a finite number of the sticks may be represented in practice, these representations yield error bounds when we truncate the outer sums in either \cref{eq:stickbreakingrep} or \cref{eq:stickbreakingrep2} to a finite number of terms \cite[\S~3.6]{doshi2009variational}; \cite[Thm.~1]{carin2011variational}; \cite[Thm.~1.5]{CUP}.
Additionally, a Markov chain Monte Carlo routine including an auxiliary variable may be used to numerically integrate over the number of represented sticks, which removes the approximation error in the asymptotic regime of the Markov chain \citep{ishwaran2001gibbs}.

\section{Controlling the statistics of latent features}
\label{sec:powerlaw}

In statistical applications, it is important to tailor the assumptions that a model encodes about the structure and complexity of the data.
In this section, we characterize the asymptotic behavior of the distribution of the latent features in the Gibbs-type IBP.
As before, let $\numfeat_n$ denote the number of dishes sampled among the first $n$ customers in the Gibbs-type IBP.
Additionally, let $\numfeat_{n,j}$ denote the number of dishes sampled exactly $j$ times among the first $n$ customers, for every $n\ge j \ge1$.

\subsection{Power-law behavior when $\discount\in (0,1)$}

As we saw in \cref{sec:partitions}, when $\discount \in (0,1)$ the underlying Gibbs-type partitions correspond to the class of partitions induced by the Poisson--Kingman measures with parameter $\tau>0$, which includes the normalized generalized gamma processes and a subclass of the Pitman--Yor processes.
These models have been shown to exhibit \emph{power-law} (i.e., heavy-tailed) behavior in the asymptotic distribution on the number of blocks in the partition \citep{pitman2003poisson}.
Empirical measurements in a variety of domains have been shown to exhibit power-law behavior.
For example, the occurrence of unique words in a document, the degrees of interactions in a protein network, and the number of citations of an academic article all exhibit power law behavior.
An appropriate model for data that may depend on these factors should be able to capture this behavior in its latent structure.
It was shown by \citet{TG2009} and \citet{BJP2012} that the Pitman--Yor IBP exhibits power-law behavior in the asymptotic distributions of $\numfeat_n$ and $\numfeat_{n,j}$.
We will now see that this behavior is, in a sense, inherited from the partitions induced by the Pitman--Yor processes, and that power-law behavior for any partition induced by a Poisson--Kingman measure translates into power-law behavior in the corresponding Gibbs-type IBP.

Let $\discount \in (0,1)$, let $\tau>0$, and let $\levygbp$ be the L\'evy intensity of the Gibbs-type beta process defined in \cref{eq:levygbp}, parameterized by the structural distribution for the Poisson--Kingman measures in \cref{eq:PKstruct}.
In this case, it follows analogously to the results by \citet[p.~459]{BJP2012} that $\levygbp$ satisfies the limiting behavior
\[
\int_{\bspace \times (0,x]} p\, \levygbp ( \dee \omega \times \dee p )
	\tends 
	\frac{\discount}{1-\discount} C x^{1-\discount}
		,
		\quad
		\text{as $x \to 0$,}
\]
for a constant $C \defas \tau^{-\discount}$, where $\tends$ indicates that the ratio of the left and right hand sides tends to one in the specified limit.
With derivations analogous to those by \citet[Prop.~6.1, Lem.~6.2, Lem.~6.3, \& Prop.~6.4]{BJP2012}, it is straightforward to verify that, with probability one,
\[
\numfeat_n &\tends \totalmass C n^\discount
	\: \text{ and } \:
\numfeat_{n,j} \tends \totalmass
	\frac{ \discount \Gamma(j-\discount) }{ j!\, \Gamma(1-\discount) }
	C n^\discount,
	\quad
	\text {as $n \to \infty$}
	,
	\label{eq:powerlaw}
\]
where $\totalmass>0$ is the mass parameter of the Gibbs-type IBP.
These statistics therefore exhibit power law behavior controlled by the value of $\discount \in (0,1)$; the closer $\discount$ is to one, the heavier the tails of these distributions.

Recall from \cref{sec:partitions} that members of the Gibbs-type partitions (and thus the Gibbs-type IBP) are obtained by mixing over a random parameter $\tau$.
Consider the case when $\tau$ has a density function on $\PosReals$ given by $h(t) \stable(t)$, for a measurable function $h \colon \PosReals \to \PosReals$.
In this case, the constant $C$ in \cref{eq:powerlaw} becomes
\[
C \defas \int_0^\infty t^{-\discount} h(t) f_\discount(t) \dee t
	.
\]
By choosing $h(t) = \frac{ \Gamma(\theta+1) }{ \Gamma(\theta/\discount + 1) }  t^{-\theta}$ for some ${\theta > -\discount}$, we have that $\levygbp$ is the L\'evy intensity of the Pitman--Yor-type beta process, and ${C=\discount^{-1} \Gamma(\theta+1) / \Gamma(\theta+\discount),}$ which was previously derived by \citet{BJP2012}.
By choosing ${h(t) = e^{ \beta^\discount - \beta t },}$ for some $\beta>0$, then $\levygbp$ is the L\'evy intensity of a normalized generalized gamma-type beta process, and we find that
$
C = e^{\beta^\discount}
	\int_0^\infty t^{-\discount} e^{-\beta t} f_\discount(t) \dee t
	.
$
In this case, if $\discount = 1/2$, then $\levygbp$ is the L\'evy intensity of a normalized inverse Gaussian-type beta process, and $C$ has a closed form solution given by
$
{C = \frac 2 \pi \beta^{1/2} e^{ \beta^{1/2} } \phi_1 ( \beta^{1/2} )} ,
$
where $\phi_\nu$ is the modified Bessel function of the third type.

\begin{figure*}
\centering
\subfigure[Asymptotic behavior of $K_n$.]{
                \includegraphics[scale=0.45]{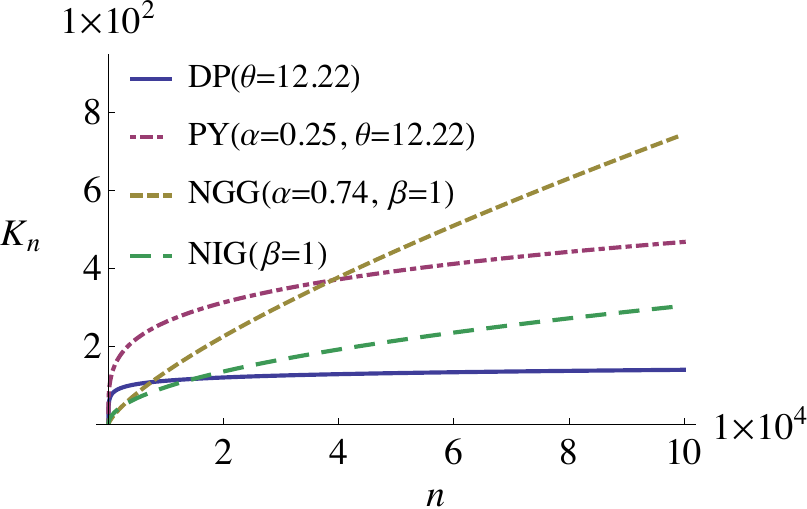}
                \label{fig:Kn}
                }
\subfigure[Asymptotic behavior of $K_{n,1}$.]{
                \includegraphics[scale=0.45]{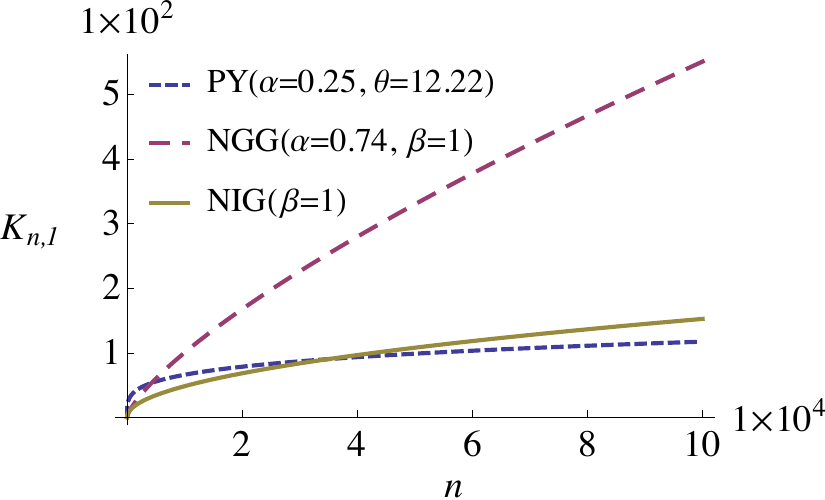}
                \label{fig:Knj}
                }
       	\caption{The behavior of $K_n$ (the number of features) and $K_{n,1}$ (the number of features with exactly one assignment) for several subclasses of the Gibbs-type IBP, as $n$ increases.  Heavy-tailed behavior demonstrates power-law properties.}
	\label{fig:powerlaw}
\end{figure*}

In order to compare the power-law behaviors of different Gibbs-type paritions, \citet{de2013gibbs} chose hyperparameters for the Pitman--Yor and normalized generalized gamma processes such that the expected number of blocks in the corresponding partitions satisfy $\EE[\numblocks_{50}] \approx 25$.
By plotting statistics such as the expected number of blocks $\numblocks_n$ in the partition as $n$ varies, one may visualize differences in the asymptotic behaviors between the models.
As one should expect, these same hyperparameter settings also provide an appropriate comparison for their corresponding Gibbs-type IBPs.
In particular, recall that in the Gibbs-type IBP the $j$-th customer samples a $\poisson( \totalmass \primitive{j-1}{1}{1} )$ number of new dishes.
Then the total number of dishes $\numfeat_n$ sampled by $n$ customers has a Poisson distribution with mean
$		\totalmass
		\sum_{j=1}^n \primitive{j-1}{1}{1},
$
where we recall that $\primitive{0}{1}{1} \defas 1$.
Setting $(\discount,\conc) = (0.25,12.22)$ and $(\discount,\beta) = (0.74,1)$ for the Pitman--Yor- and normalized generalized gamma-type IBPs, respectively, we then have that $\EE[\numfeat_{50}] \approx 25 \gamma$ for both models.
In \cref{fig:powerlaw}, we plot the behavior of $\numfeat_n$ and $\numfeat_{n,1}$ as $n$ increases for these two Gibbs-type IBP subclasses, with the additional choice of $\totalmass=1$.

We can see that, for this comparable set of hyperparameters, the normalized generalized gamma-type IBP exhibits heavier tails than the Pitman--Yor-type IBP on both statistics, though in smaller $n$ regimes the reverse holds.  The normalized inverse Gaussian-type IBP, at the same setting of $\beta=1$, exhibits similar tail behavior in $\numfeat_{n,1}$ to the Pitman--Yor-type IBP.  For comparison, the asymptotic behavior of $\numfeat_n$ for the Dirichlet-type IBP at the same hyperparameter setting as the Pitman--Yor-type IBP is also displayed, which does not exhibit power-law behavior ($\numfeat_n$ grows proportionally with $\log n$ in this case \citep{GGS2007}). 
These characteristics distinguish the subclasses of Poisson--Kingman-type IBPs and provide a variety of power-law modeling options to a practitioner.
%
%

\subsection{Logarithmic growth when $\discount = 0$}

Recall that the Gibbs-type partitions with $\discount = 0$ coincide with the random partitions induced by the Dirichlet processes with concentration parameter $\theta$. 
With probability one, the number of blocks in the partition of $[n]$ satisfies $\numblocks_n \sim \theta \log n$ as $n\to \infty$ \citep{korwar1973contributions}.
Similarly, with probability one, the number of features in the corresponding Gibbs-type IBP (i.e., the original IBP) 
satisfies $\numfeat_n \sim \totalmass \theta \log n$ as $n\to \infty$ \citep{GGS2007}, 
where $\gamma$ is the mass parameter of the IBP.

\subsection{Finite feature models when $\discount < 0$}
\label{sec:finitemodel}

Finally, recall that the Gibbs-type partitions with $\discount <0$ coincide with the random partitions induced by the Pitman--Yor processes with discount parameter $\discount < 0$ and concentration parameter $\theta = m \vert \discount \vert$ (see \cref{eq:pyparams}), where $m$ is a \emph{random} element in $\Nats$ \citep[Chp.~3, Sec.~2]{pitman2002combinatorial}; \citep[Thm.~12]{gnedin2006gibbs}.
This subclass may be interpreted as an urn scheme with a finite---but random---number of colors $m$, and a number of specific examples have been investigated in the literature \citep{gnedin2010species,de2013gibbs}.
In this case, with probability one, $\numblocks_n = m$ for all sufficiently large $n$.
That is, there are a finite number of blocks that are eventually exhausted.

As one may anticipate, the corresponding Gibbs-type IBP in this regime may be analogously interpreted as a feature allocation with a random finite number of features.
In particular, when $\discount<0$, the Gibbs-type IBP will have a finite number of features if and only if $\EE[m] < \infty$.
Informally, recall from \cref{sec:bepbp} that the number of new dishes $\numfeat_{n+1}^+$ sampled by the $n+1$-st customer in the Gibbs-type IBP is Poisson distributed with rate $\totalmass \Pr \event{ \numblocks_{n+1} > \numblocks_n }$ (see \cref{eq:ordinaryZ,eq:structspecies}). Using two Borel--Cantelli arguments, we show in the supplementary material that $\EE[m] < \infty$ if and only if the sequence $\Pr \event{ \numblocks_{n+1} > \numblocks_n }$ is summable if and only if $\numfeat_{n+1}^+ = 0$ for all sufficiently large $n$ a.s.

\section{Black-box posterior inference}
\label{sec:inference}

We propose a Markov chain Monte Carlo algorithm generalizing the procedure for posterior inference with the IBP, originally developed by \citet{GGS2007} and \citet{meeds2007modeling}.
We will see how these inference methods may be treated as a black-box, where implementing any subclass of the Gibbs-type IBP requires only several evaluations of the primitives $\primitive{n}{z_1}{z_2}$, given by \cref{eq:primitives}.

Fix $n\ge 1$,
and let $(\omega_1, \dotsc, \omega_{\numfeat_n})$ denote the $\numfeat_n$ distinct atoms among the sequence $Z_1, \dotsc, Z_n$, where, in this section, we assume that the ordering is chosen uniformly at random, conditioned on $\numfeat_n$.  
For every $i \le n$ and $k\le \numfeat_n$, define $Z_{i,k} \defas Z_i (\theset{\bomega_k})$, 
and let $Z \defas \gprocess{Z_{i,k}}{i\le n,\, k\le \numfeat_n}$.
Latent feature models have been applied to a variety of statistical problems (as discussed in \cref{sec:ibp}).
In most of these applications, the features (associated with the atoms) represent latent clusters or factors underlying a data set comprised of $n$ observations $Y \defas (Y_1, \dotsc, Y_n)$.
Informally,  observation $Y_i$ is associated with every latent component $\omega_k$ for which $Z_{i,k} = 1 $.
More carefully, 
let $\Omega = (\bomega_1,\dotsc,\bomega_{\numfeat_n})$ and recall that $\Omega$ is an i.i.d.\ sequence (drawn from the normalized base measure) and independent of $Z$, conditioned on $\numfeat_n$.
Let $\psi$ be a latent variable independent of $\Omega$ and $Z$, and define $\Phi = (\psi,\Omega)$.
We then fix a likelihood $p ( Y | Z, \Phi ) = \prod_{i=1}^n f(Y_i ; \psi, Z_i)$ for some density $f$.
In other words, the numbering of the features is irrelevant to the likelihood.

Consider resampling an element of $Z$ from its conditional distribution given $Y$, $\Phi$, and the remainder of $Z$.
Fix a data point $i \le n$. 
For every $k \le \numfeat_n$, let $Z_{-(i,k)}$ be the elements of $Z$ excluding $Z_{i,k}$,
let $Z^z_{-(i,k)}$ be the elements of $Z$ with $Z_{i,k}$ replaced by $z$,
and let $\smash{S_k^{(-i)}} \defas \sum_{j\ne i, j \le n} Z_{j,k}$ be the number of datapoints, other than $i$, that have feature $k$.
For $k \le \numfeat_n$ and $S_k^{(-i)} >0$, 
Bayes's rule implies that 
\[
\label{eq:posteriorZ}
\begin{split}
&\Pr \cevent{ Z_{i,k} = z \given Y, Z_{-(i,k)} , \Phi }
	\propto p ( Y \given Z^z_{-(i,k)}, \Phi )
		\times
		\Pr \cevent{ Z_{i,k} = z \given Z_{-(i,k)} }
		,
		\quad z \in \{0, 1\}
		,
\end{split}
\]
where $p( Y \given Z, \Phi )$ is the likelihood defined above.
Recall that we have associated the $i$-th customer in the buffet analogy with $Z_i$.
By exchangeability, we may treat this as the last customer to enter the buffet, and so
\[\label{esthiuhtde}
\Pr \cevent{ Z_{i,k} = 1 \given Z_{-(i,k)} }
	\propto
		(S_k^{(-i)} - \discount)
		\primitive{n-1}{1}{0}.
\]
Therefore, conditioned on $\numfeat_n$, for every $i \le n$ and $k \le \numfeat_n$ where $S_k^{(-i)}>0$,
we may resample $Z_{i,k}$ according to \cref{eq:posteriorZ,esthiuhtde}.

We can resample the remaining elements of $Z$ using the Metropolis--Hastings proposal proposed by \citet{meeds2007modeling}.
In particular, for every $i \le n$, we propose removing those features possessed by only $Z_i$, that is, those atoms $\bomega_k$ in $\{\bomega_1, \dotsc, \bomega_{\numfeat_n}\}$ with $Z_{i,k} =1$ and $S_k^{(-i)}=0$.
We propose replacing these atoms with $\numfeat_i^+$ new atoms (possessed only by $Z_i$).
Recall that $\numfeat_i^+$ is interpreted as the number of dishes taken by only the $i$-th customer.
Because we may treat the $i$-th customer as if they were the last to enter the buffet, the distribution of $\numfeat_i^+$ is the same as the distribution of the number of new dishes sampled by the last customer, and so 
\[
\label{eq:newfeatsampling}
\numfeat_i^+ \dist \poisson(\totalmass \primitive{n-1}{1}{1})
	.
\]
The Markov proposal replaces those entries in $\Phi$ associated with the removed atoms with a set of new parameters associated with the new atoms, 
sampled from the normalized base measure. (In order to get a simple acceptance probability, the numbering of the features, and thus the column ordering of the array Z, can be resampled uniformly at random. Alternatively, one can ignore the ordering and work implicitly in the space of equivalence classes up to ordering, as the columns are already uniquely identified by their latent parameters, assuming the base measure is non-atomic.)
Let $Z^*$ and $\Phi^*$ denote the proposed feature assignments and parameters.
It is straightforward to show that the Metropolis--Hastings acceptance probability for this proposal is
{$
\min \{
	1 , p(Y\vert Z^*, \Phi^*)/p(Y\vert Z, \Phi)
	\}
$}
 \citep{meeds2007modeling}.
This move potentially changes the number of atoms $\numfeat_n$ among $Z_1, \dotsc, Z_n$ and thus the number of latent features in the feature allocation.
We then proceed to the next process $Z_{i+1}$ and repeat this procedure.
Iterating these steps, along with standard Gibbs sampling moves that resample the latent parameters $\Phi$, results in a Markov chain that targets the posterior distribution of $Z$ and $\Phi$, conditioned on the data $Y$, as its steady state distribution.

Without good prior knowledge of what the parameters $\totalmass$, $\discount$ and $\Theta$ governing the IBP model should be for a particular application and data set, we may give them broad prior distributions and infer their values during posterior inference.
See \cref{sec:experiments} for further details. 
Note that the inference procedure we have described may be treated as a black-box for any subclass of Gibbs-type IBPs, where the user only needs to supply several evaluations of the primitives $\primitive{n}{\cdot}{\cdot}$.
In particular, resampling $Z$ only requires the two values $\primitive{n-1}{1}{1}$ and $\primitive{n-1}{1}{0}$ (in order to evaluate \cref{eq:posteriorZ,eq:newfeatsampling}) for a dataset of size $n$.
In order to resample the hyperparameters $\totalmass$, $\discount$ and $\Theta$ for the IBP model, one needs to supply $n-1$ additional evaluations to obtain $\primitive{n-s}{s}{1}$, for $n\ge s\ge 1$, required by \cref{eq:pmf}.
These $n+1$ values may be precomputed and stored for given values of $\discount$ and $\Theta$. 
See the supplementary material for some notes on computing these primitives, the required generalized factorial coefficients $\GFC(n,k;\discount)$ in \cref{eq:gfc}, and the Gibbs-type weights $\Varray$ for different models.

\section{Experiments}
\label{sec:experiments}

We now demonstrate the differences between several subclasses of the Gibbs-type IBP. 
We do not implement models with $\discount<0$ here due to computational difficulties (as discussed in the supplementary material).
This section will therefore focus on subclasses of the Gibbs-type IBP with $\discount \in [0,1)$.  
See \cref{sec:conclusion} for a further discussion.

For every $i\le n$, assume that data point $Y_i$ is composed of $p$ measurements \: \: ${Y_i \defas (Y_{i,1}, \dotsc, Y_{i,p})}$.
Consider the following factor analysis model for $Y$:
\[
\label{eq:datamodel}
Y_{i,j} = \sum_{k=1}^{K_n} W_{i,k} Z_{i,k} A_{k,j} + \varepsilon_{i,j}
	,
	\qquad  i\le n,\, j\le p
	,
\]
where $W \defas (W_{i,k})_{k\le K_n, i\le n}$ are $\Reals$-valued modulating weights,
$A \defas (A_{k,j})_{k\le K_n, j\le p}$ are $\Reals$-valued factor loadings, and $\varepsilon \defas (\varepsilon_{i,j})_{j\le p, i\le n}$ are $\Reals$-valued additive noise terms.
Let
\[
&W_{i,k} \given \sigma_W \dist \Normal (0, \sigma_W^2)
	,
	&&\quad i\le n,\, k\le K_n
	,
	\label{eq:latentfactormodel1}
	\\
&A_{k,j} \given \sigma_{A,j} \dist \Normal(0, \sigma_{A,d}^2)
	,
	&&\quad j\le p,\, k\le K_n
	,
	\label{eq:latentfactormodel2}
	\\
&\varepsilon_{i,j} \given \sigma_Y \dist \Normal(0, \sigma_Y^2)
	,
	&&\quad i\le n,\, j\le p
	,
	\label{eq:latentfactormodel3}
\]
where $\sigma_Y, \sigma_W, \sigma_{A,1}, \dotsc, \sigma_{A,p}$ are positive-valued hyperparameters. 
Viewing $Y$, $Z$, $W$, $A$, and $\varepsilon$ as matrices in the obvious way, we may write $Y = (W \circ Z) A + \varepsilon$ where $\circ$ represents element-wise multiplication.
Then the data $Y$ is conditionally matrix Gaussian and admits the conditional density
\[
\label{eq:likelihood}
p(Y \given Z, W, A, \sigma_X)
	= 
	\frac 1 {(2\pi)^{np/2} \sigma_X^{np}}
		\exp \Bigl \{
			-\frac 1 {2\sigma_X^2}
			\text{tr} \Bigl [
				(Y - M)^T (Y-M)
			\Bigr ]
		\Bigr \}
	,
\]
where $M= (W \circ Z) A$.
Note that, in practice, $W$ or $A$ may be analytically marginalized out of this likelihood expression, in which case $Y$ is still conditionally Gaussian.

In the experiments below, we give all hyperparameters broad prior distributions and resample their values during inference with \emph{slice sampling} \citep{Neal2003}.
Where relevant, the discount parameter $\discount$ is given a $\betadist(1,1)$ prior distribution.
All other parameters in $\Theta$ (i.e., the Gibbs-type hyperparameters, which are all positive-valued) are given independent gamma prior distributions, whose hyperparameters are themselves given independent $\expdist(1)$ prior distributions.
For the noise parameter $\sigma_Y$, we let $\sigma_Y^{-2} \given a_Y, b_Y \dist \gammadist(a_Y, b_Y)$, where $a_Y, b_Y \dist \expdist(1)$ are independent.
We give $\sigma_W$ a similar prior specification.
Independently of $\sigma_Y$ and $\sigma_W$, we couple the factor variance parameters $(\sigma_{A,j})_{j\le p}$ with a similar model: let $\sigma_{A,j}^{-2} \given a_A, b_A \dist \gammadist(a_A, b_A)$, for all $j\le p$, where $a_A, b_A \dist \expdist(1)$ are independent.
Finally, for the IBP mass parameter, let $\totalmass \given a_\totalmass, b_\totalmass \dist \gammadist(a_\totalmass, b_\totalmass)$, and let $a_\gamma, b_\gamma \dist \expdist(1)$ be independent.
In this case,  \cref{eq:pmf} implies the conditional distribution of $\totalmass$ remains in the family of gamma distributions, with conditional density 
\[
p( \totalmass \given Z, \discount, \Theta, a_\gamma, b_\gamma )
	&\propto
		\totalmass^{\numfeat_n}
			 \exp \Bigl (
			 	 -\totalmass 
				 \sum_{j=1}^n \primitive{j-1}{1}{1} 
				\Bigr )
				\times
				\gammadist(\totalmass; a_\gamma, b_\gamma)
				\\
		&= \gammadist \Bigl (
			 	 \totalmass ;
				 a_\gamma + \numfeat_n
				 ,
				 b_\gamma + \sum_{j=1}^n \primitive{j-1}{1}{1} 
				\Bigr )
	.
\]

\subsection{Synthetic data}

\begin{figure*}
\centering
\subfigure[Latent feature matrix.]{
	\includegraphics[scale=0.4]{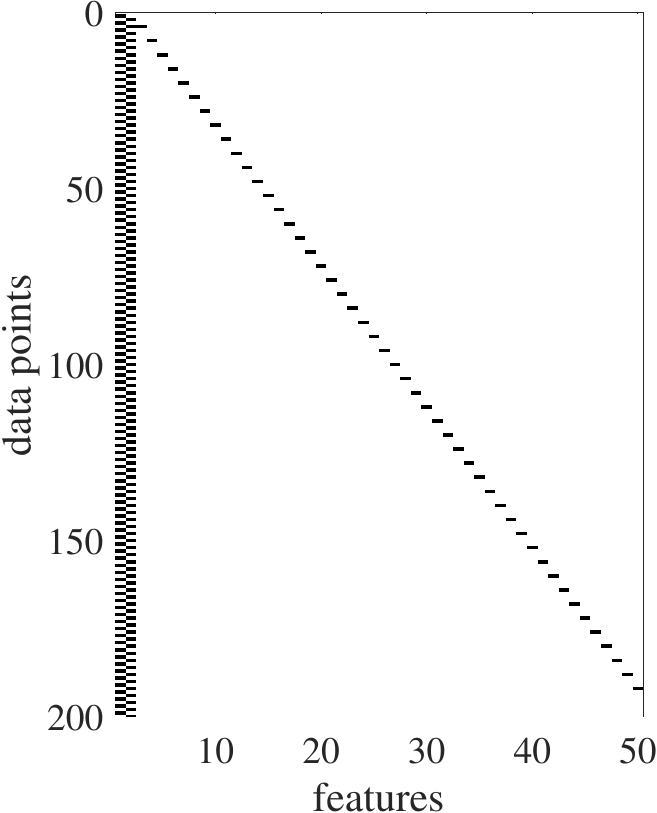}
	\label{fig:syntheticZ}
	}
	\hfill
\subfigure[Inferred number of features.]{
	\includegraphics[scale=0.4]{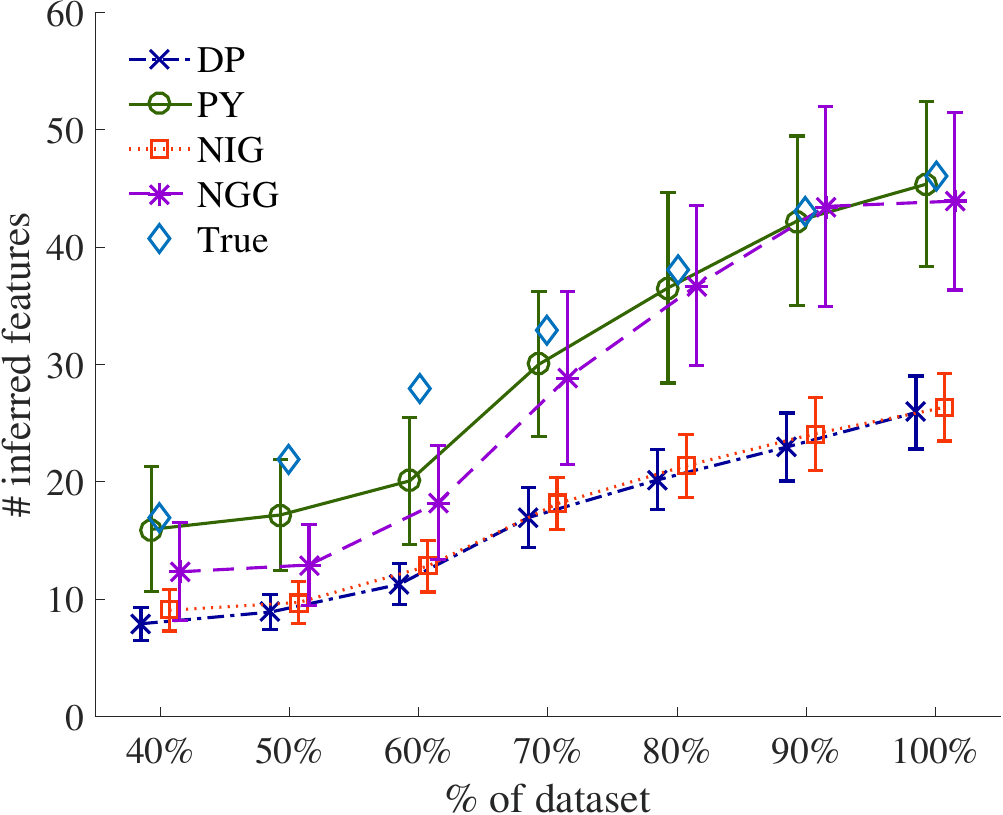}
	\label{fig:syntheticK}
	}
\caption{(a) A synthetic latent feature matrix for $n=200$ data points with $\numfeat_{200}=50$ features.  The simulated data was in $p=50$ variables.  (b) The number of features inferred by different subclasses of the Gibbs-type IBP as we sequentially include more of the data.  For each subset of the data, we plot the mean number of features over 3,000 samples following a burn-in period.  Bars at $\pm$ one standard deviation are also displayed.  The true number of features in each subset of the data is plotted for reference.}
\label{fig:synthetic_experiment}
\end{figure*}

First consider a synthetic latent feature allocation, displayed as a $200\times 50$ binary matrix in \cref{fig:syntheticZ}.
The rows correspond to the $n=200$ data points and the columns correspond to the $\numfeat_n=50$ latent features, that is, the $i$-th row and $k$-th column is shaded black if $Z_{i,k}=1$ (in the notation of \cref{sec:inference}).
In this example, every data point possesses one of the first two features, and the remaining 48 features are each only possessed by one data point.
We simulate a dataset $Y$ of $n=200$ measurements in $p=50$ variables from the model in \cref{eq:datamodel,eq:latentfactormodel1,eq:latentfactormodel2,eq:latentfactormodel3} with $\sigma_X = \sigma_W = 1$, and $\sigma_{A,j} = 1$ for $j\le p$.

We implemented the posterior inference procedure described in \cref{sec:inference} for 6,000 burn-in iterations.
In \cref{fig:syntheticK} we display the number of features inferred by the Dirichlet, Pitman--Yor, normalized inverse Gaussian, and normalized inverse gamma---denoted DP, PY, NIG, and NGG, respectively---subclasses of the Gibbs-type IBP on different subsets of the data.
In particular, we ran the inference procedure on 40\% of the data points, then on 50\%, and so on, indicated by the horizontal axis from left to right.
The mean number of inferred features (along with $\pm$ one standard deviation) over 3,000 samples following the burn-in period are displayed for each model.
The true number of features in each subset of the data are also displayed for reference.

We note that all models attained approximately the same training loglikelihood given each data subset (averaged over the samples).
However, the more flexible PY and NGG-IBP variants were able to more accurately infer the number of features underlying the data compared to the less expressive subclasses, the DP- and NIG-IBPs.
We recall that the DP-IBP is an extreme point of both the PY- and NGG-IBP subclasses.
The discount parameter $\discount$ differentiates these models, and as we saw in \cref{sec:powerlaw}, inferring this parameter allows these models to detect the power law structure present in the latent feature allocation displayed in \cref{fig:syntheticZ}.
In the supplementary material, we provide trace plots of the Gibbs-type hyperparameters over the burn-in period, along with histograms over samples repeatedly drawn following the burn-in.

%
%

\subsection{MNIST digits}

\begin{figure*}
\centering
\subfigure[DP; \# images sharing each feature]{
	\includegraphics[scale=0.36]{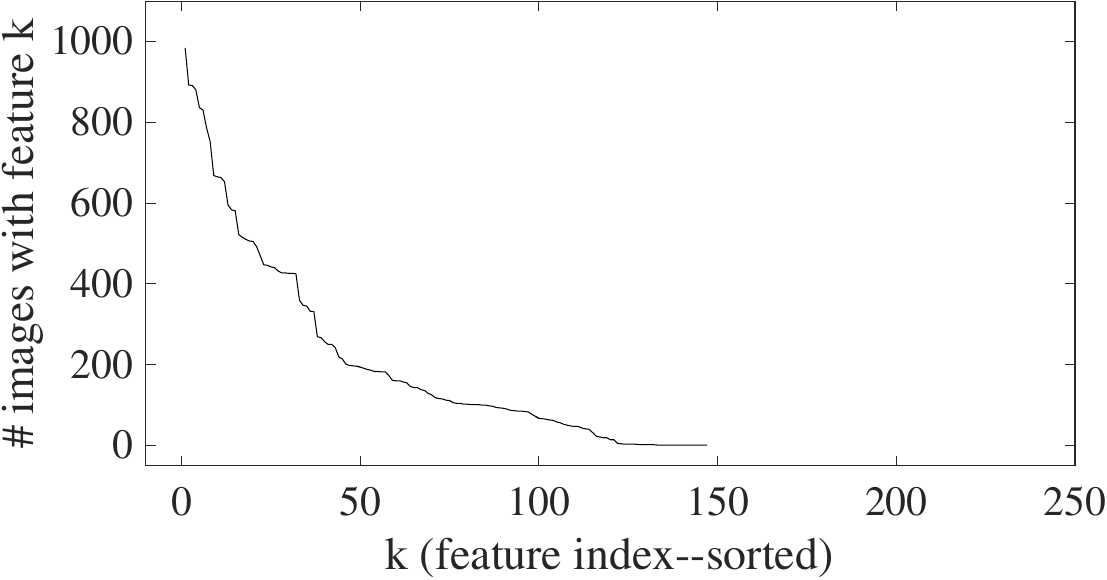}
	}
	\hfill
\subfigure[DP; \# feat. used by each image]{
	\includegraphics[scale=0.25]{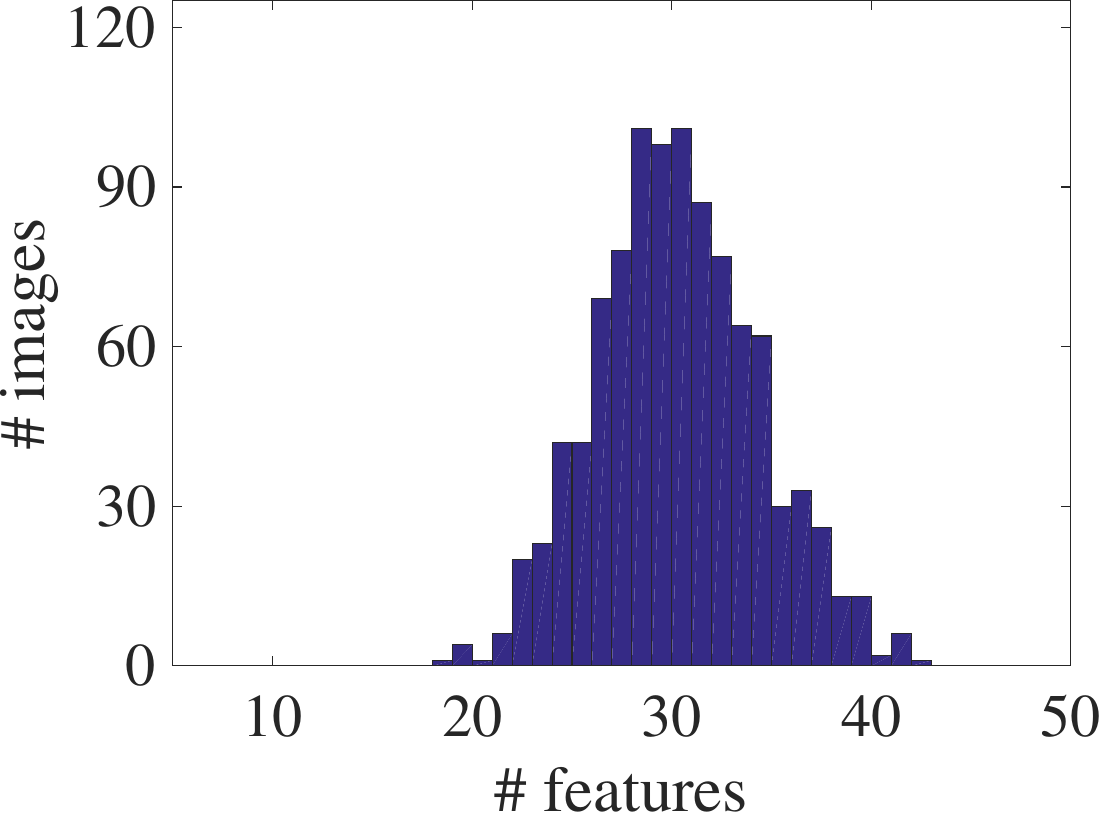}
	}
\subfigure[PY; \# images sharing each feature]{
	\includegraphics[scale=0.36]{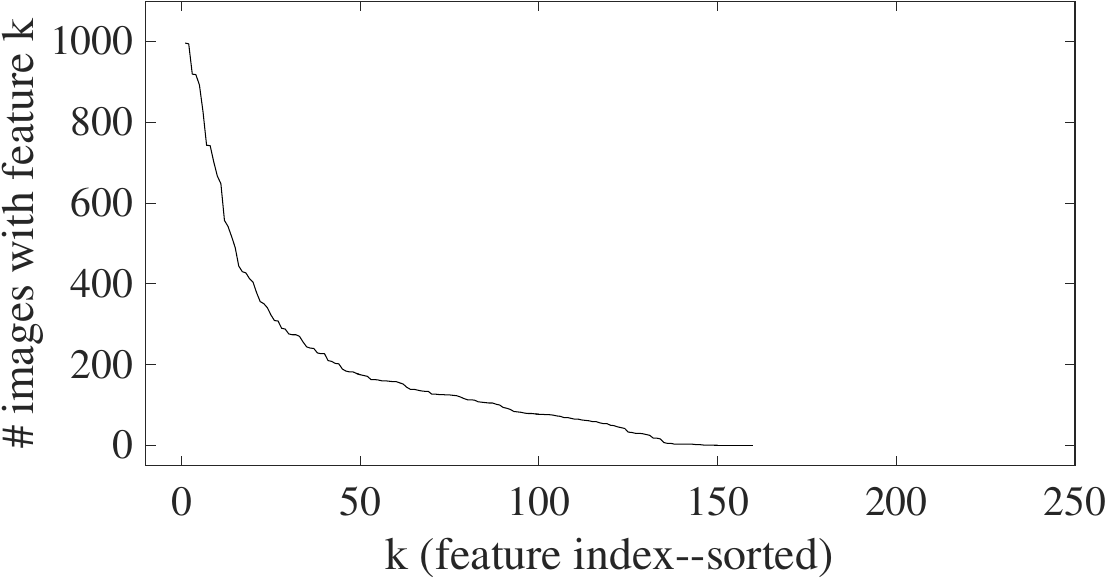}
	}
	\hfill
\subfigure[PY; \# feat. used by each image]{
	\includegraphics[scale=0.25]{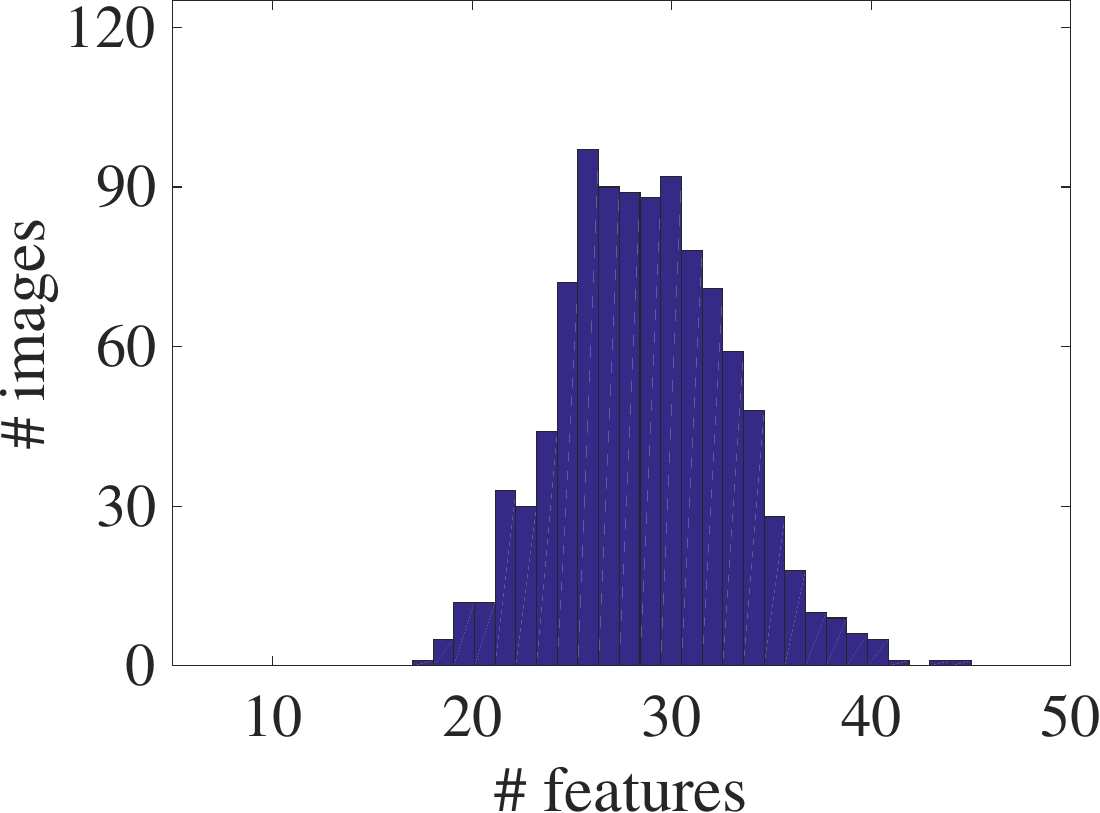}
	}
\subfigure[NIG; \# images sharing each feature]{
	\includegraphics[scale=0.36]{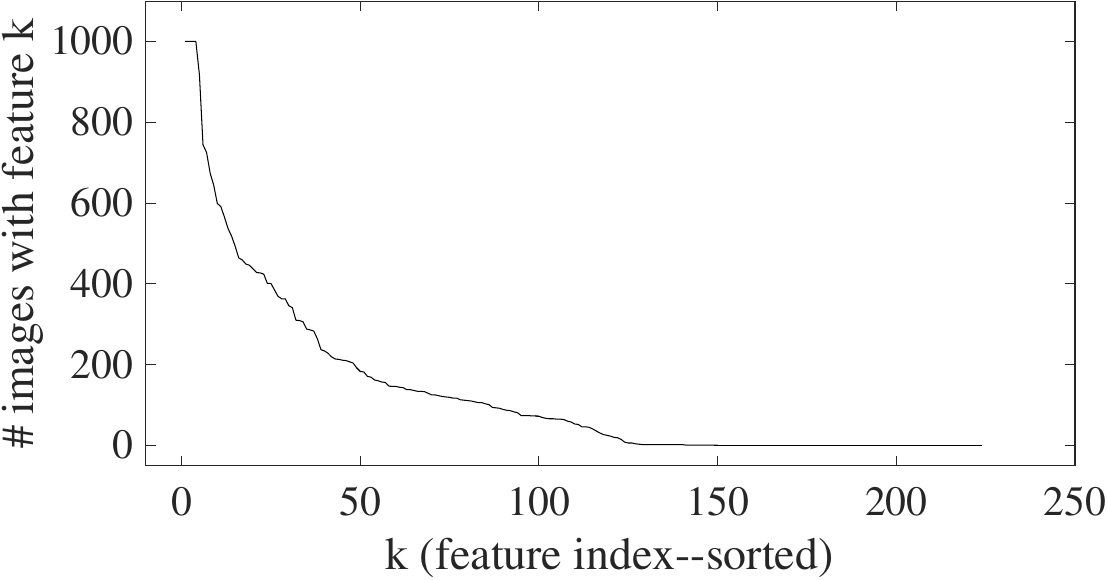}
	}
	\hfill
\subfigure[NIG; \# feat. used by each image]{
	\includegraphics[scale=0.25]{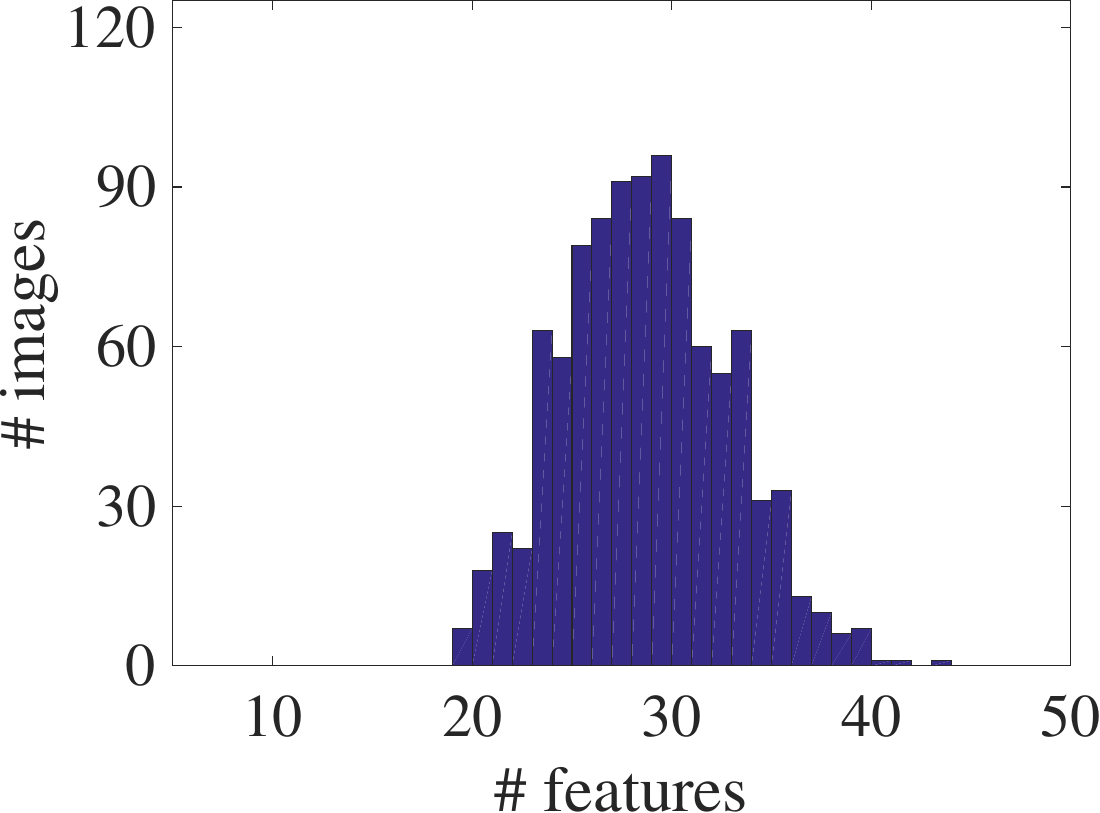}
	}
\subfigure[NGG; \# images sharing each feature]{
	\includegraphics[scale=0.36]{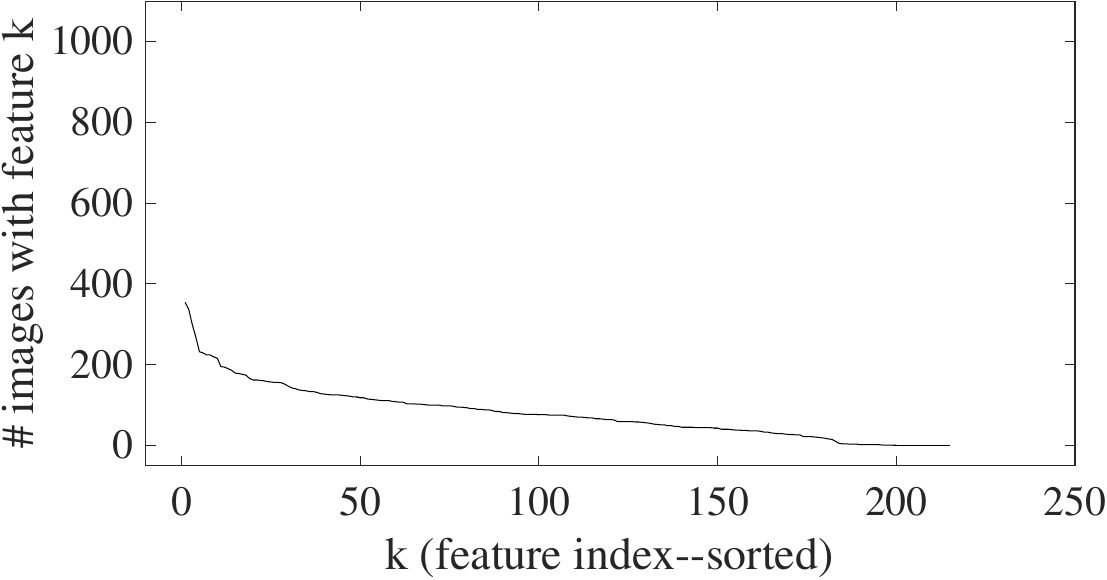}
	}
	\hfill
\subfigure[NGG; \# feat. used by each image]{
	\includegraphics[scale=0.25]{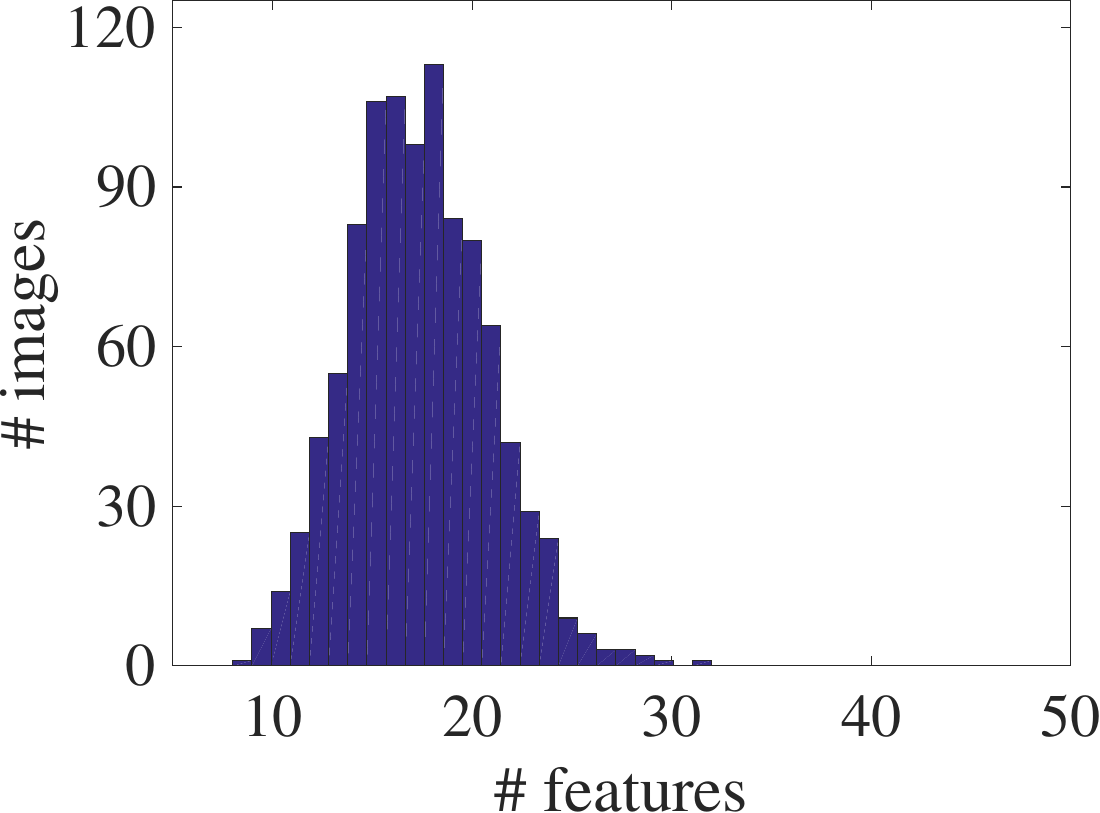}
	}
\caption{Latent feature statistics inferred by each model on the MNIST dataset.  For each model, the number of images assigned to each feature is displayed as a plot (sorted for visualization), and the number of features used by an image is displayed as a histogram.}
\label{fig:mnist_feat_stats}
\end{figure*}

We also applied the model in \cref{sec:inference} to $n=1000$ examples of the digit `3' from the MNIST handwritten digits dataset.
We projected the data onto its first $p=64$ principal components in order to replicate the experiment performed by \citet{TGG07} with the DP-IBP (and a more restrictive setting of the hyperparameters).
Here we present the same qualitative analyses for different subclasses of the Gibbs-type IBP.
The reader can see \cite{PZWGC2010,BJP2012} for similar experiments.
We ran our posterior inference procedure for 20,000 iterations, which was sufficient for every model to burn-in.
We collected 1,000 samples (thinned from 10,000 samples) of all latent variables in the model following the burn-in period, and we display boxplots of the number of inferred features over the collected samples in \cref{fig:mnist_num_feat}.
(In the supplementary material, we provide visualizations of the inferred values of the Gibbs-type hyperparameters.) 
In \cref{fig:mnist_feat_stats}, we find the MAP sample (of all latent variables and parameters, from among the collected samples) for each model, and for that sample we plot (1) the number of images sharing each feature and (2) a histogram of the number of features used by each image.
For visualization, the features in the former plots are ordered according to the number of images assigned to them.
The scale of the axes in the subfigures are held fixed for comparison.

\begin{figure*}
\centering
\includegraphics[scale=0.4]{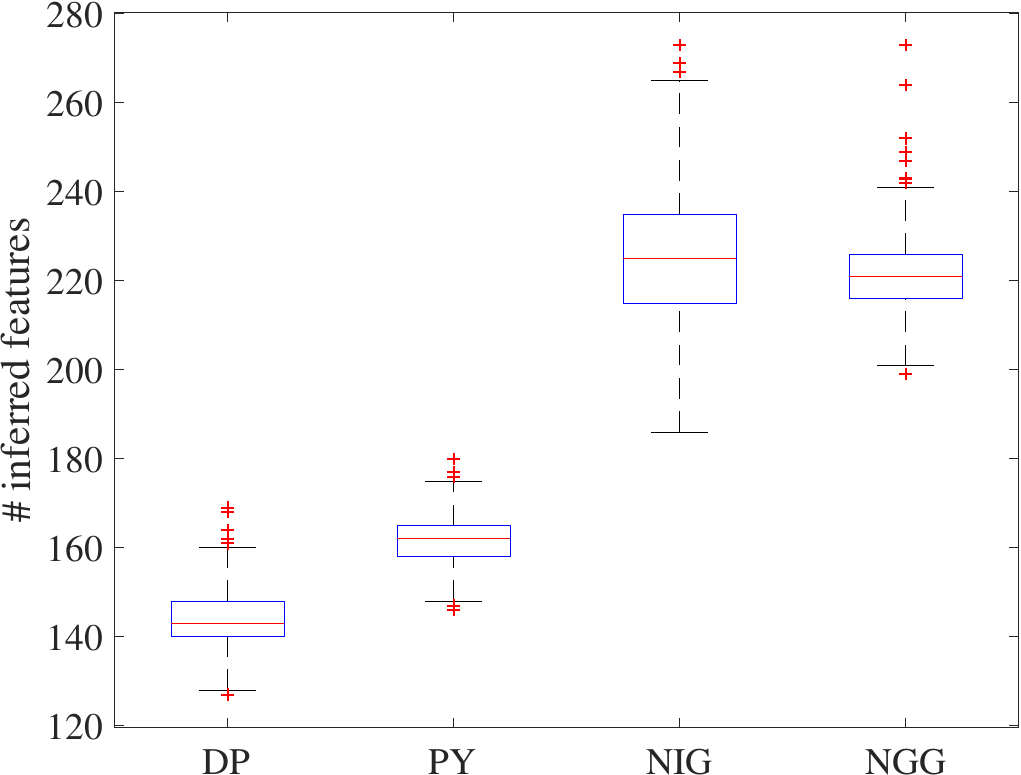}
\caption{Number of features inferred by the different subclasses on the MNIST dataset.  Boxplots over 1,000 samples (thinned from 10,000 samples) collected following a burn-in period of 20,000 iterations.}
\label{fig:mnist_num_feat}
\end{figure*}

We find that the heavy-tailed models, i.e., the PY-, NIG-, and NGG-IBPs, exhibit different extents of power-law behaviors achieved by tailoring the total number of inferred features and the number of features with relatively few assignments.
In particular, \cref{fig:mnist_num_feat} shows that the PY-IBP infers more features than the DP-IBP (based on an unpaired t-test at a 0.05 significance level).
Moreover, both the NIG- and NGG-IBP models infer significantly higher numbers of features than the PY-IBP, but do not themselves differ significantly.
\cref{fig:mnist_feat_stats} shows that these differences are due to varying power-law behaviors between the models.
In particular, the PY-, NIG-, and NGG-IBP models display increasingly heavier tail behavior in the (distribution of the) number of images sharing each feature.  
The NGG-IBP model is notable as clearly having dramatically heavier tails than all other models in this distribution.
This additionally results in a noticeably lower average number of features per image (visible in the histogram), which does not appear to differ significantly between the other three subclasses.

This experiment demonstrates important variations between the Gibbs-type IBP subclasses.
Compare the latent feature distributions between the three heavy-tailed variants.
On one hand, the NIG-IBP has heavier tails than the PY-IBP, accomplished by creating many features to which very few images are assigned, resulting in a significantly larger number of features.
On the other hand, the NGG-IBP has much heavier tails than the NIG-IBP, accomplished by heavily skewing the distribution towards the (right) tail, yet maintaining approximately the same total number of features.
It is particularly interesting to compare the PY- and NGG-IBP models in this respect, as the DP-IBP may be approximated by both of these subclasses.
As discussed in \cref{sec:powerlaw}, these differing properties 
provide several different options to a practitioner, which are accessible through our black-box constructions and posterior inference procedures.

\begin{figure*}
\centering
\begin{tabular}{c}
DP-IBP
	\\
\includegraphics{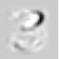}
\includegraphics{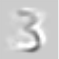}
\includegraphics{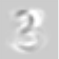}
\includegraphics{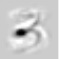}
\includegraphics{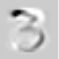}
\includegraphics{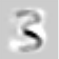}
\includegraphics{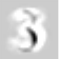}
\includegraphics{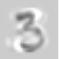}
\includegraphics{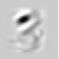}
\includegraphics{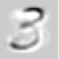}
	\\
PY-IBP
	\\
\includegraphics{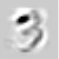}
\includegraphics{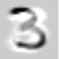}
\includegraphics{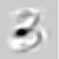}
\includegraphics{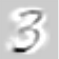}
\includegraphics{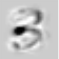}
\includegraphics{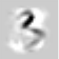}
\includegraphics{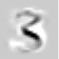}
\includegraphics{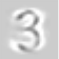}
\includegraphics{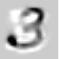}
\includegraphics{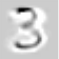}
	\\
NIG-IBP
	\\
\includegraphics{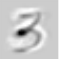}
\includegraphics{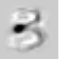}
\includegraphics{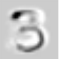}
\includegraphics{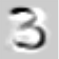}
\includegraphics{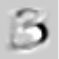}
\includegraphics{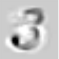}
\includegraphics{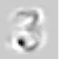}
\includegraphics{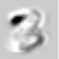}
\includegraphics{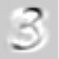}
\includegraphics{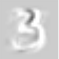}
	\\
NGG-IBP
	\\
\includegraphics{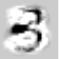}
\includegraphics{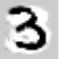}
\includegraphics{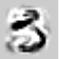}
\includegraphics{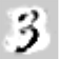}
\includegraphics{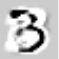}
\includegraphics{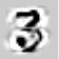}
\includegraphics{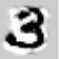}
\includegraphics{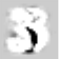}
\includegraphics{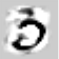}
\includegraphics{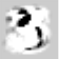}
\end{tabular}
\caption{Top 10 (according to the weight matrix $W$) important features (represented by the factors in $A$) for the digit `3' inferred by each subclass of the Gibbs-type IBP.  Darker pixel values correspond to larger values in (the corresponding factor in) $A$.}
\label{fig:mnist}
\end{figure*}

Finally, we can visualize the effect that the different latent feature distributions have on this particular application by investigating some of the latent features inferred by each model.
In \cref{fig:mnist}, we display the top 10 (according to the weight matrix $W$) most important features (represented by the factors in $A$) from the MAP sample collected for each model.
The features inferred by the DP-, PY-, and NIG-IBP models do not appear to differ, however, the NGG-IBP clearly places the heaviest weight on its features (resulting in darker pixel values).
Moreover, a few of these features appear to capture distinct parts of the digits.

%
%

\section{Conclusion}
\label{sec:conclusion}

The Gibbs-type IBPs are a broad class of feature allocation models, parameterized by the law of a Gibbs-type random partition.
We showed how the Gibbs-type IBP can be constructed from exchangeable sequences of completely random measures and gave several stick-breaking representations.
We also characterized the asymptotic behavior of the number of latent features in a Gibbs-type IBP, which was seen to mimic the asymptotic behavior of the underlying random partition.
We described black-box routines for simulation and performing posterior inference with Gibbs-type IBPs that only require a set of precomputed constants that are specific to the corresponding partition law.
\enlargethispage{.25\baselineskip}
Our numerical experiments demonstrated differences between the Gibbs-type IBP subclasses, where we saw that different extents of heavy tailed latent feature behavior could be attained beyond the PY-IBP.

Many models that use the beta process as a basic building block can be generalized by instead using the Gibbs-type beta process, which could benefit many applications of the IBP. 
Further applications of the beta process beyond the IBP should also be considered.
For example, \citet{CUP} provides a finitary construction for exchangeable sequences of Bernoulli processes (as in \cref{eq:gibbsbernoulliprocess}) rendered conditionally \iid\ by a \emph{hierarchical beta process} \citep{TJ2007}.
Such processes are used as admixture models, in which a collection of feature allocations share features, analogously to (collections of) random partitions induced by a hierarchy of partitioning schemes. 
Feature allocations induced by hierarchies of Gibbs-type beta processes would be a natural generalization of this framework, providing flexible properties (such as power law behavior) to the admixture model.

Finally, we cannot practically apply the simulation or inference procedures described in this article to Gibbs-type IBPs for $\discount<0$, because we cannot robustly compute the required primitives $\primitive{n}{\cdot}{\cdot}$ in this case (as described in the supplementary material).
Constructions by \citet[Def.~6.1]{CUP} provide alternative simulation procedures, however, posterior inference algorithms have yet to be developed.
The stick-breaking representations in \cref{sec:stickbreaking} do not depend on these primitives, and so they may suggest an approach for inference.

\section*{Supplementary Material}

Supplementary Material: Gibbs-type Indian buffet processes\\ 
(DOI: \href{10.1214/19-BA1166SUPP}{https://doi.org/10.1214/19-BA1166SUPP}).

\bibliographystyle{ba}
\bibliography{gibbs-type-ibp}

\begin{acknowledgement}
We thank Stefano Favaro for help with (and code for) implementing various Gibbs-type models and for communicating most of the material in Sec.~S.2 of the supplementary material to us.
We thank Lancelot F. James and Igor Pr{\"u}nster for helpful advice and pointers to references.
Finally, we thank Annalisa Cerquetti, Hong Ge, Konstantina Palla, Christian Steinruecken, and anonymous reviewers for feedback on drafts.
CH was supported in part by the Stephen Thomas studentship at Queens' College, Cambridge, with funding also from the Cambridge Trusts.
DMH was supported in part while a research fellow of Emmanuel College, Cambridge, with funding also from a Newton International Fellowship and AFOSR grant \#FA9550-15-1-0074.
\end{acknowledgement}

\end{document}